\documentclass{article}
\PassOptionsToPackage{table}{xcolor}
\usepackage{xcolor}
 \usepackage{preprint,times}

\usepackage{amsmath,amsfonts,bm}

\def\eqref#1{equation~\ref{#1}}

\def\1{\bm{1}}

\DeclareMathAlphabet{\mathsfit}{\encodingdefault}{\sfdefault}{m}{sl}
\SetMathAlphabet{\mathsfit}{bold}{\encodingdefault}{\sfdefault}{bx}{n}

\usepackage[bookmarks=true,colorlinks=true,citecolor=blue,linkcolor=blue,urlcolor=blue]{hyperref}
\usepackage{url}
\usepackage{tcolorbox}

\usepackage{xspace}

\usepackage{graphicx}
\usepackage{subcaption}

\usepackage{wrapfig}
\usepackage{needspace}
\usepackage{amsmath}
\usepackage{amssymb}
\usepackage{booktabs,multirow,tabularx}
\usepackage{algorithm}
\usepackage{algpseudocode}
\usepackage[normalem]{ulem}

\makeatletter
\renewcommand{\paragraph}{\@startsection{paragraph}{4}{\z@}{0.1ex \@plus .1ex}{-1em}{\normalfont\normalsize\bfseries}}
\makeatother

\definecolor{bestbg}{HTML}{d6e2e9}
\definecolor{secondbg}{HTML}{edeec9}

\newcommand{\besthl}[1]{{\setlength{\fboxsep}{1.2pt}\colorbox{bestbg}{\textbf{#1}}}}
\newcommand{\secondhl}[1]{{\setlength{\fboxsep}{1.2pt}\colorbox{secondbg}{\textbf{#1}}}}

\usepackage{array}

\newcolumntype{B}{>{\columncolor{gray!5}}c}  \newcolumntype{O}{>{\columncolor{gray!8}}c}

\title{Flow-based Policy Adaptation\\ without Policy Updates}

\newcommand{\preprintauthornote}{\begingroup
  \normalfont\fontsize{8}{10}\selectfont
  \raggedright
  \setlength{\parskip}{0pt}Equal contribution.
  Emails:
  \href{mailto:luzhesun@ttic.edu}{\texttt{luzhesun@ttic.edu}},
  \href{mailto:jijingtian@ttic.edu}{\texttt{jijingtian@ttic.edu}},
  \href{mailto:haoran.chen@ttic.edu}{\texttt{haoran.chen@ttic.edu}},
  \href{mailto:jiawei.zhou.1@stonybrook.edu}{\texttt{jiawei.zhou.1@stonybrook.edu}},
  \href{mailto:mwalter@ttic.edu}{\texttt{mwalter@ttic.edu}}.\par
  \endgroup
}
\author{Luzhe~Sun\textsuperscript{1,}\thanks{\protect\preprintauthornote}\quad
Jingtian~Ji\textsuperscript{1,}\footnotemark[1]\quad
Haoran~Chen\textsuperscript{1}\quad
Jiawei~Zhou\textsuperscript{2}\quad
Matthew~R.~Walter\textsuperscript{1}}
\newcommand{\preprintaffiliations}{\textsuperscript{1}Toyota Technological Institute at Chicago\qquad
\textsuperscript{2}Stony Brook University}

\newcommand{\alg}{{\normalfont\scshape Gloves}\xspace}

\usepackage[nohyperlinks]{acronym}
\acrodef{GRIP}[{\normalfont\scshape Grip}]{Guided Refinement with an Intent-Preserving Energy}
\acrodef{LIFT}[{\normalfont\scshape Lift}]{Learned Inversion-Free Transport}
\acrodef{VLA}{vision-language-action}
\acrodefplural{VLA}{vision-language-action models}
\acrodef{CFG}{classifier-free guidance}
\acrodef{DiT}{Diffusion Transformer}
\newcommand{\partialflow}{\ac{GRIP}\xspace}
\newcommand{\flowalign}{\ac{LIFT}\xspace}

\makeatletter
\renewcommand{\@maketitle}{\begingroup
    \centering
    {\normalfont\fontsize{18}{22}\selectfont\bfseries\@title\par}
    \vspace{0.9em}
    {\normalfont\fontsize{10}{12}\selectfont\mbox{\@author}\par}
    \vspace{0.4em}
    {\normalfont\fontsize{9}{11}\selectfont\preprintaffiliations\par}
  \endgroup
  \vspace{0.8em}
}
\makeatother

\begin{document}

\maketitle

\begin{abstract}
Foundation models and other policies trained on large and diverse datasets can endow robots with semantic and task knowledge that would be difficult to acquire from task-specific demonstrations alone.
Despite their broad scene understanding and reasoning capabilities, however, these models often lack the environment- and embodiment-specific knowledge required for precise and reliable execution.
We propose \alg, a flow-based framework that complements an existing agent with a lightweight, goal-agnostic expert action prior that refines the action chunks proposed by the agent without updating its parameters.
The prior captures observation-conditioned expert behavior, while the agent's proposal supplies task- and context-specific intent that the correction aims to preserve.
The learned flow also provides a natural in-distribution scoring mechanism through reverse flow evaluation. We use this signal as an intervention gate: actions that appear consistent with the expert distribution are passed through unchanged, while anomalous or out-of-distribution (OOD) actions are corrected. In this way, \alg only provides assistance when necessary. By learning local goal-agnostic expert action patterns and using them to adapt agent-proposed actions during execution, \alg provides a lightweight shared-control module for robust action adaptation across tasks and environments.
Across controlled OOD-recovery studies, simulation benchmarks, and real-robot experiments, \alg consistently improves the success of \acp{VLA} and Code-as-Policy agents without retraining the underlying policy.
Code and demos are available at \href{https://ripl.github.io/GLOVES_web}{\nolinkurl{ripl.github.io/GLOVES_web}}.

\end{abstract}
\acresetall

{

\begin{center}
    \includegraphics[width=\linewidth]{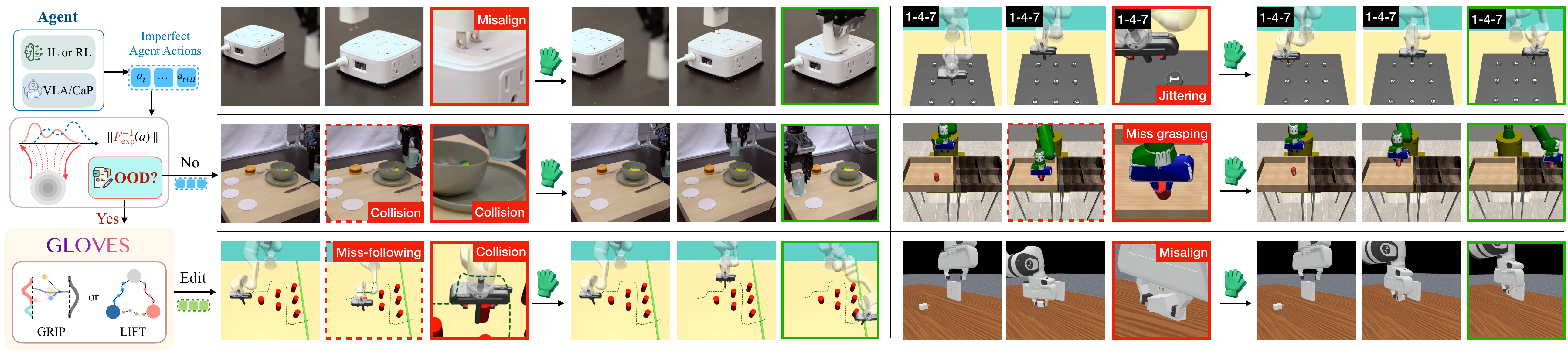}
    \captionof{figure}{\small Overview of \alg. At inference time, an
    agent-proposed action chunk is passed to the OOD gate to determine
    whether the chunk should be executed directly or adapted. When
    intervention is needed, \alg refines the action with \acs{GRIP} or \acs{LIFT}
    and applies the refined chunk for the robot's rollout.}
    \label{fig:teaser}
\end{center}

\section{Introduction}
Modern robotic systems increasingly rely on large pretrained policies that bring broad prior knowledge to new tasks, particularly \ac{VLA} models trained on large and diverse datasets. With knowledge of the goal or intent (e.g., conveyed via language), these models provide high-level reasoning capabilities that would be difficult to acquire from task-specific demonstrations alone.
Yet these capabilities alone are not sufficient to ensure reliable execution---the resulting actions are often noisy, biased, delayed, or invalid due to domain shift~\citep{jones2025beyond}. 
Typically, these errors are addressed by finetuning the policy using additional task- and environment-specific demonstrations~\citep{kim2024openvla,kim2025openvla-oft,team2024octo,myers2024policy}.
However, finetuning can be prohibitively expensive or simply not possible when the agent is only accessible via a black-box API.
Even when the parameters can be updated, incorporating task-specific expertise directly into the policy requires modifying a model whose broad pretrained capabilities we may wish to preserve.
We therefore ask: \emph{Can an existing agent's execution be improved through action-level correction, 
without any policy updates?}

Viewed through the lens of shared autonomy, we formulate this problem as image-conditioned action-chunk adaptation~\citep{Zhao2023ACT, Chi2023DiffusionPolicy, Janner2022Diffuser} and propose \alg (Figure~\ref{fig:teaser}), a flow-based framework for adapting an existing agent's actions.
\alg uses goal-agnostic expert demonstrations to learn a lightweight, image-conditioned expert action prior. Critically, unlike the agent, which is conditioned on the task goal (e.g., specified via language), \alg does not assume access to the goal during action adaptation.
Because the same observation can admit multiple valid behaviors, the agent's proposed chunk provides an essential cue to the goal-dependent behavior it intends to execute~\citep{yoneda2023noise, sun2025flashback,sun2025stackgen}.
\alg refines this proposal using the learned prior while regularizing toward the original chunk.

\alg performs this adaptation through a generative action editing process, using the expert prior to steer the agent's proposed action toward expert behavior while preserving the agent's intent. Balancing these potentially competing objectives is non-trivial. Existing generative editing methods typically sample actions through a noising-denoising process~\citep{Meng2022SDEdit, yoneda2023noise}, where the noise level controls the trade-off between the \textit{fidelity} of the resulting action and its \textit{conformity} to expert behavior.
\alg provides two strategies for dealing with this tradeoff. \partialflow combines partial noising with energy-guided generation, requiring only an expert flow. \flowalign additionally learns an agent flow and applies FlowAlign-style source-to-expert transport~\citep{Lipman2023FlowMatching, Liu2023RectifiedFlow,Kim2025FlowAlign} without selecting a partial-noising depth. Both strategies regularize the correction toward the agent's original proposal.

\alg uses the reverse direction of the trained flow model as an in-distribution scoring mechanism and only adapts actions that are identified as anomalous or out-of-distribution (OOD). Action chunks that pass through this \emph{flow-based OOD gate} are executed unchanged. The same expert flow thus supports both OOD detection and action correction, without training a separate detector. In contrast, existing methods~\citep{yoneda2023noise,reddy2018shared,sun2025flashback} either adapt \emph{every} action chunk, at the cost of repeated ODE integrations and potentially overriding valid agent behavior, or they require training a separate uncertainty model in order to decide when to provide assistance~\citep{mcmahan2024shared}.

Our contributions are threefold:
(1) \textbf{Goal-agnostic action adaptation without policy updates.}
We introduce \alg, a lightweight framework that combines an
expert action prior with agent-proposed chunks through
\partialflow{} and \flowalign{}.
The same expert flow supports selective intervention and action
correction without a separately trained detector. (2) \textbf{Improved execution for VLA and CaP agents.}
We demonstrate action adaptation for both vision-language-action
(VLA) and code-as-policy (CaP) agents across three simulation
tasks while keeping the underlying agents unchanged.
On Slalom, \flowalign raises VLA success from $33.06\%$
to $76.67\%$ and CaP success from $63.67\%$ to $97.67\%$.
Real-robot VLA experiments demonstrate complementary capabilities:
fine-grained correction for precise charger insertion and
sustained trajectory refinement for smooth cup transport.
On charger insertion, \partialflow{} increases success from
$45\%$ ($18/40$) to $100\%$ ($40/40$). (3) \textbf{Controlled analysis of detection and correction.}We complement end-to-end evaluation with structured perturbations that provide known injection events for statistical analysis of OOD detection and task-performance recovery across error types. Comparisons with and without gating, together with computational profiling, characterize the benefits and costs of selective
intervention.

\section{Related Work}
\paragraph{Generative Models for Imitation Learning}
Generative policies are well suited for imitation learning, where demonstrations can exhibit multimodal action distributions when tasks can be performed in different ways. Prior work models this multimodality using implicit discrete sequence models~\citep{Shafiullah2022BehaviorTransformer, Zhao2023ACT}. 
Diffusion models formulate planning and control as conditional denoising, enabling trajectory-level behavior synthesis~\citep{Janner2022Diffuser}, offline decision making~\citep{Ajay2023DecisionDiffuser}, and visuomotor imitation~\citep{Reuss2023BESO, Chi2023DiffusionPolicy, Ze2024DP3}.

The iterative sampling required by diffusion policies can be costly for closed-loop control, motivating single- or few-step distillation methods such as Consistency Policy~\citep{Prasad2024ConsistencyPolicy} and One-Step Diffusion Policy~\citep{Wang2025OneDP}. 
Flow-matching policies offer an alternative by learning a velocity field that maps a source distribution to the target distribution, 
with applications to robot motion learning, manipulation, and generalist robot policies~\citep{Braun2024RFMP, Zhang2025FlowPolicy, Jung2025CoFPolicy, Black2025PiZero, Yan2025ManiFlow}. 
Whereas these methods primarily use generative models to synthesize actions from noise, \alg instead adapts agent-proposed actions, selectively moving them toward the expert distribution, while preserving the agent’s intent.

\paragraph{Action Adaptation}

Policy adaptation aims to improve pretrained robot policies (e.g., those trained via imitation learning (IL)
or Reinforcement Learning (RL)), including \ac{VLA} models~\citep{kim2024openvla, kim2025openvla-oft, pan2026sop, myers2024policy}. Some methods directly finetune \acp{VLA}~\citep{collaboration2023open, wang2024scaling, team2024octo, Black2025PiZero} or adapt them through language-level decomposition~\citep{myers2024policy, shi2024yay},
while others add failure prediction~\citep{Xu2025FailDetect}, corrective supervision~\citep{yang2026fpc}, or online post-training~\citep{xu2024rldg} to improve deployment performance. Closest to our setting are residual or correction-based methods that keep the base policy intact and learn an auxiliary module to refine its proposed actions~\citep{yuan2024policy, welte2026flowcorrect}. Related, generative shared-autonomy methods use diffusion~\citep{yoneda2023noise} and consistency models~\citep{sun2025flashback} to refine actions proposed by a human rather than a pretrained policy while seeking to preserve intent. Selective-intervention methods further study when assistance should be applied~\citep{mcmahan2024shared, yu2025armada}, reduce unnecessary assistance, only intervene when the copilot is expected to improve performance.
Our approach combines expert-prior-guided editing of action chunks with selective intervention: the agent's proposed action chunk conveys intent, while the same expert flow supports both OOD scoring and editing.

\section{Problem Formulation}
\label{sec:problem_formulation}

Let 
$I$ denote the image observation, and $g$ the task goal.
Given the goal, an existing agent proposes an action chunk $\operatorname{Agent}: (I,g)\mapsto a_u \in \mathbb{R}^{T\times d_a}$, where $T$ is the chunk length and $d_a$ the action dimension. The adapter receives the proposed action chunk $a_u$ and image observation $I$.
It produces a refined chunk according to
\begin{equation}
    \tilde a \sim \mathcal{A}_\theta(\cdot\mid a_u,I).
\end{equation}
We learn an expert action prior from context-action pairs
$\mathcal{D}_e=\{(I_i,a_i^e)\}$. Explicit \textbf{goal} conditioning is \textbf{excluded} from both prior learning and action correction, although the demonstrations may be collected by a goal-informed expert.
The agent's proposal therefore serves as a cue to the intended behavior. Our objective is to improve task execution while preserving this intent, while not modifying the underlying agent. We exclude explicit goal conditioning from both prior learning and action correction to decouple execution refinement from the agent's goal representation, which may take the form of language, target states, or images. This allows the expert prior to support agents with different goal interfaces without requiring a shared goal encoder or goal-labeled correction data. The base agent remains responsible for interpreting the task, while its proposed action chunk supplies the adapter with an intent cue.

\subsection{Preliminary: Flow Matching}
\label{sec: preliminary}
Flow matching~\citep{Lipman2023FlowMatching, Kim2025FlowAlign} is a generative modeling framework that learns to draw samples from a target data distribution \(\mathcal{P}_1\) by continuously transporting samples drawn from a tractable source distribution \(\mathcal{P}_0\) (e.g., a standard Gaussian). 
Concretely, it transports a sample \(x_0\sim\mathcal{P}_0\) to a sample \(x_1\sim\mathcal{P}_1\) by solving the ODE
\begin{equation}
    \frac{d x_t}{dt} = u_t(x_t), \quad t \in [0,1].
\end{equation}
Flow matching learns a neural approximation \(v_\theta(x,t)\) for the true velocity field \(u_t(x_t)\) parameterized by \(\theta\).
The ideal flow-matching objective regresses the model prediction to the ground-truth velocity \(u_t(x)\) of a probability path \(\mathcal{P}_t\) connecting \(\mathcal{P}_0\) and \(\mathcal{P}_1\):
\begin{equation}
    \min_\theta \; 
    \mathrm{E}_{t\sim\mathcal U(0,1),\,x_t\sim\mathcal P_t}
    \left[ \| v_\theta(x_t,t) - u_t(x_t) \|_2^2 \right].
\end{equation}
Since \(u_t\) is generally intractable~\citep{Lipman2023FlowMatching},
conditional flow matching instead supervises the model with velocities from conditional paths between paired or coupled samples \((x_0,x_1)\). In the rectified-flow setting~\citep{Liu2023RectifiedFlow}, the conditional path is the linear interpolation
\begin{equation}
    x_t = (1-t)x_0 + t x_1,
    \qquad
    u_t(x_t \mid x_0,x_1) = x_1 - x_0.
\end{equation}
Training then reduces to solving a supervised regression problem over sampled times and pairs,
\begin{equation}
    \min_\theta \; \mathrm{E}_{t,x_0,x_1}
    \left[ \| v_\theta((1-t)x_0 + t x_1,t) - (x_1-x_0) \|_2^2 \right].
\end{equation}

Given an external condition \(c\), such as an observation, the target distribution becomes \(\mathcal{P}_1(\cdot\mid c)\), and the learned velocity field is written as \(v_\theta(x,t\mid c)\). The training objective has the same form as above, with the expectation taken over \(c\) and samples from the corresponding conditional path. Integrating $v_\theta(\cdot,t\mid c)$ from $t=0$ to $t=1$ defines the conditional flow map $F_\theta(\cdot;c)$.

\section{When to adapt: Out-of-Distribution Detection}

\begin{wrapfigure}[15]{r}{0.33\textwidth}
    \vspace{-\baselineskip}
    \centering
    \includegraphics[width=\linewidth]{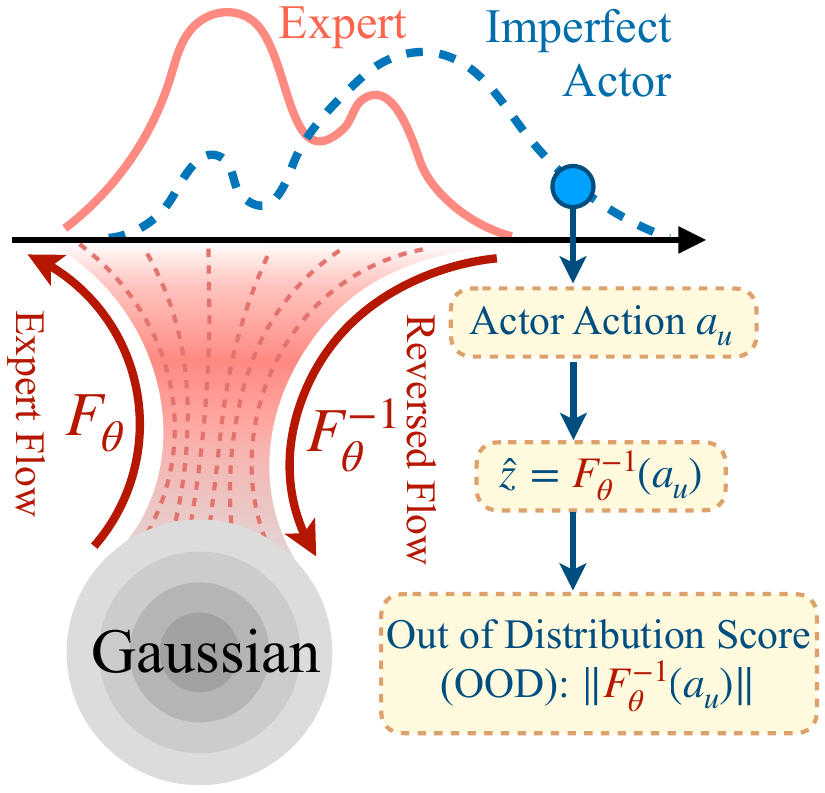}
    \caption{\small OOD detection.}
    \label{fig:ood-detection}
    \vspace{-\baselineskip}
\end{wrapfigure}

We use the reverse velocity of the trained expert flow to construct a lightweight out-of-distribution (OOD) score, inspired by flow-based failure detection~\citep{Xu2025FailDetect}.

Given an agent-proposed action chunk $a_u$ and the corresponding image observation $I_u$,
the reverse flow $F_\theta^{-1}(a_u;I_u)$ provides a latent representation associated with the Gaussian base distribution. We approximate this representation from flow time $t=1$ (target) to $t=0$ (Gaussian), and define the nonconformity score $s$ as
\begin{equation}
        \hat z(a_u;I_u) = a_u-v_\theta(a_u,1\mid I_u), \qquad s(a_u;I_u) = \|\hat z(a_u;I_u)\|_2^2.
    \label{eq:reverse_flow_score}
\end{equation}
Larger scores are treated as larger nonconformity to the expert distribution. To test whether action $a_u$ is OOD,
we calibrate this score using held-out expert examples
$\mathcal{D}_{\mathrm{cal}}=\{(a_i,I_i)\}_{i=1}^{n}$.
Writing $s_i=s(a_i;I_i)$, (notably, $I_u$ does not have to be from $\{I_i\}$)
we compute the conformal $p$-value
\begin{equation}
    p(a_u;I_u)
    =
    \frac{
        1+\sum_{i=1}^{n}
        \mathbf{1}\{s_i\geq s(a_u, I_u)\}
    }{n+1}.
    \label{eq:ood_pvalue}
\end{equation}
A small $p$-value indicates that the proposed chunk has an unusually large score relative to the expert reference data. When OOD gating is enabled, we apply action correction if $p(a_u;I_u) \leq \alpha$ and otherwise execute the proposed chunk unchanged. The threshold $\alpha$ controls gating sensitivity, with larger
values triggering more frequent intervention. We precompute expert calibration scores before deployment, so online gating requires only a single- or few-step reverse-flow evaluation and comparison against the cached scores, supporting real-time intervention decisions. As a check on the score itself, using it alone to guide a local search over action chunks improves task success on all three simulation tasks (Appendix~\ref{parag: local search}, Table~\ref{tab:ood_local_search}).

\begin{figure}[!t]
      \centering
        \begin{subfigure}[b]{0.45\textwidth}
        \centering
        \includegraphics[width=\textwidth, height=5cm, keepaspectratio]{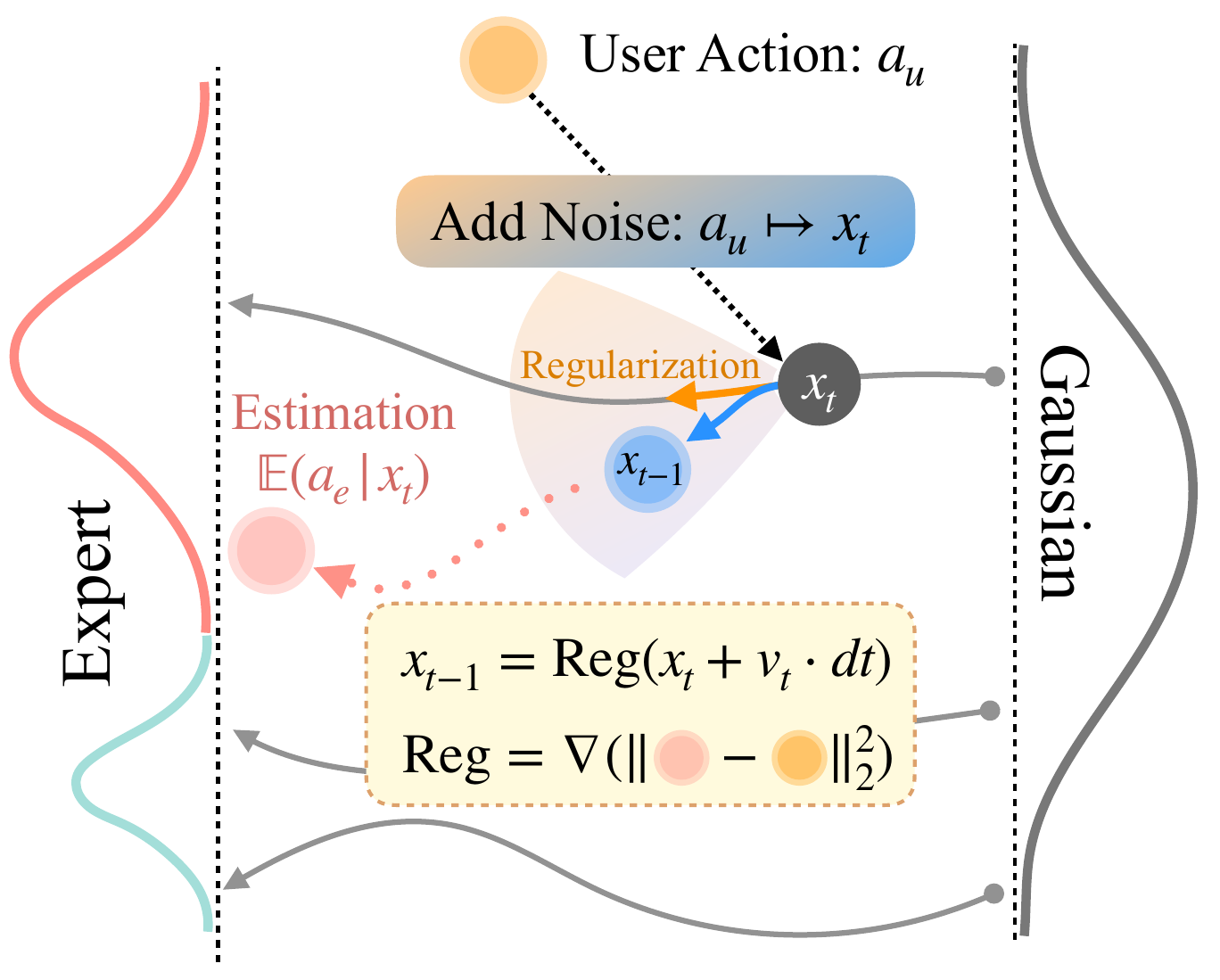}
        \caption{\acs{GRIP}}
        \label{fig:FEEG}
      \end{subfigure}
      \hfill
      \begin{subfigure}[b]{0.45\textwidth}
        \centering
        \includegraphics[width=\textwidth, height=5cm, keepaspectratio]{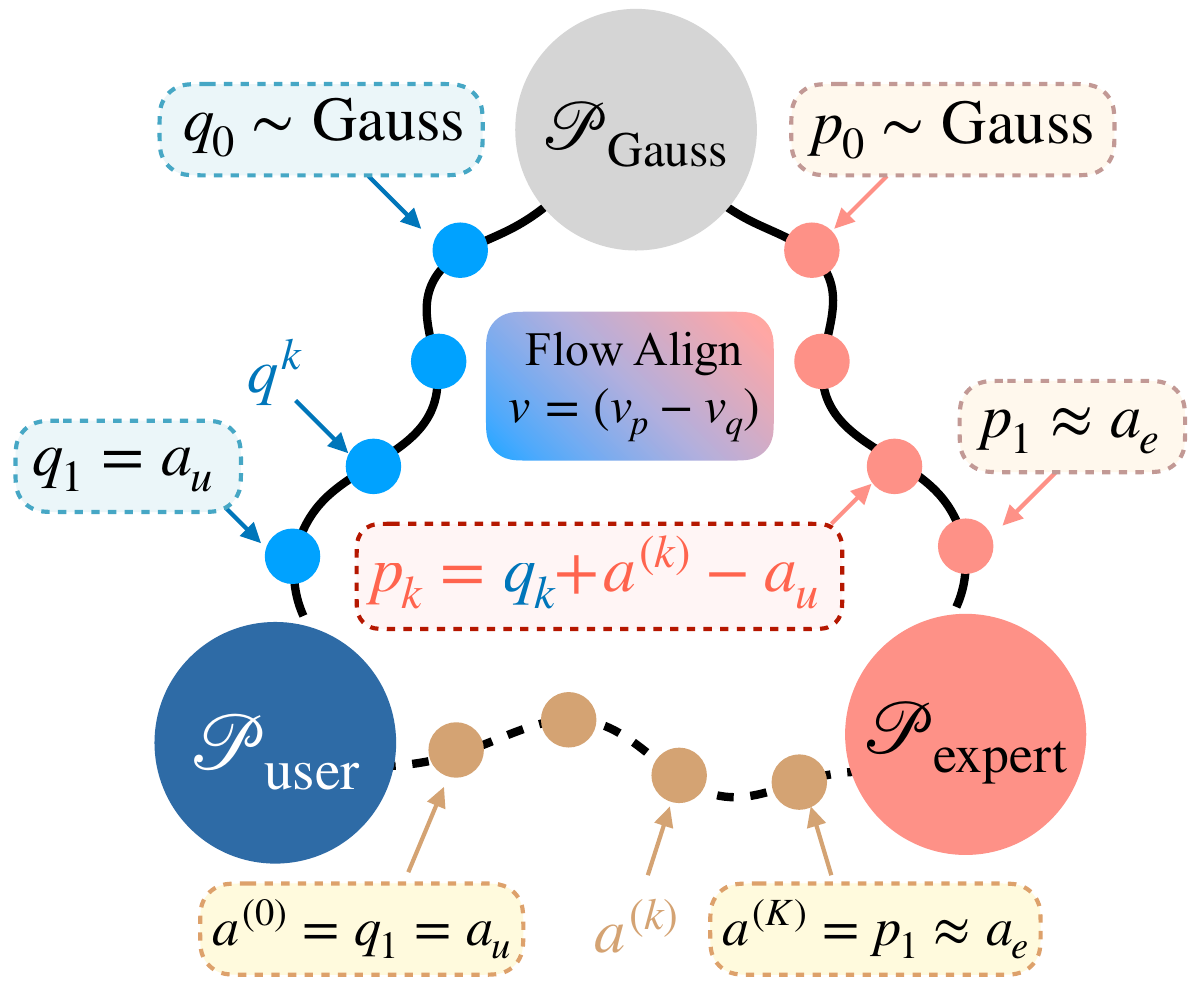}
        \caption{\acs{LIFT}}
        \label{fig:flowalign-detail}
      \end{subfigure}
      \caption{\small Overview of the policy adaptation methods in \alg. \subref{fig:FEEG} \acs{GRIP} follows the expert flow while regularizing toward the original agent command. \subref{fig:flowalign-detail} \acs{LIFT} uses both agent and expert flows to transport agent actions toward the expert distribution without explicit inversion.}
      \vspace{-5mm}
      \label{fig:all}
\end{figure}

\section{How to adapt: Gloves}
\subsection{\acs{GRIP}: Single-Flow Action Editing}
\label{sec: subsec: FSA}
Inspired by inversion-based sample editing~\citep{yoneda2023noise, sun2025flashback, Meng2022SDEdit}, we present \textbf{\acl{GRIP} (\acs{GRIP})}, which directly guides the learned conditional expert flow to generate expert-like actions that are consistent with the agent's intent. \partialflow optionally adopts OOD detection to skip editing when the agent's action is already in the expert distribution. Specifically, during inference, we initialize the trajectory from a noisy version of \(a_u\)
\begin{equation}
    x_{t_{\mathrm{init}}}
    = t_{\mathrm{init}}a_u+(1-t_{\mathrm{init}})\epsilon,
    \qquad \epsilon\sim\mathcal{N}(0,\mathbf{I}),
\end{equation}
where \(t_{\mathrm{init}}\in[0,1]\) controls the autonomy level. With larger values, \ac{GRIP} adheres more closely to the agent's actions,
whereas smaller values resample more aggressively from the expert prior. 
To further preserve the agent's intent, we use energy-guided flow matching~\citep{Feng2025FlowGuidance}, with an energy function that penalizes deviation from the agent's original proposed action:
\begin{equation}
    E(x_1) = \frac{1}{2}\left\|x_1-a_u\right\|_2^2 .
\end{equation}
During integration, \acs{GRIP} evaluates this energy function using a one-step Euler estimate $\hat{x}_1$ of the terminal action $x_1$ under the expert flow:
\begin{equation}
  \hat{x}_1(x_t,t) = x_t+(1-t)v_\theta(x_t,t\mid I),
\end{equation}
\acs{GRIP} calculates the refined action chunk $\tilde a$ by integrating the guided velocity field $\tilde{v}$ from $t = t_{\mathrm{init}}$ to $t = 1$:
\begin{equation}
    \tilde v(x_t,t\mid I)
    =
    v_\theta(x_t,t\mid I)
    -\lambda_{\text{align}} \nabla_{x_t}E(\hat{x}_1(x_t,t)),
\end{equation}
where \(\lambda_{\text{align}}\ge 0\) controls guidance strength. For \partialflow, across all experiments, we find that $\lambda_{\text{align}} =1$ yields stable and effective performance. 

\subsection{\acs{LIFT}: Dual-Flow Action Editing}
\label{sec:flowalign_editor}

\textbf{\acf{LIFT}} extends the expert-only formulation by additionally learning a flow from agent rollouts.
Following FlowAlign~\citep{kulikov2025flowedit,Kim2025FlowAlign}, it edits action chunks using the difference between agent and expert-guided velocity fields, with a source-consistency regularizer based on their estimated clean endpoints. This replaces partial noising with source-to-expert transport and removes the need to choose a partial-noising depth.

Let $v_u$ and $v_e\equiv v_\theta$ denote the image-conditioned velocity fields learned from agent rollouts ($v_u$) and expert demonstrations ($v_e$), respectively (Figure~\ref{fig:flowalign-detail}).
We maintain an editing state $a^{(k)}$ on the time grid $0=t_0<\cdots<t_K=1$(similar to local search in sec \ref{parag: local search}), initialized at $a^{(0)}=a_u$.
At editing step $k=0,\ldots,K-1$, we sample \(\epsilon_k\sim\mathcal{N}(0,I)\) and construct the agent intermediate sample $q_k = (1-t_k)\epsilon_k + t_k a_u$. We then construct the corresponding expert intermediate sample by shifting \(q_k\) according to the current editing state: $p_k = q_k + \left(a^{(k)} - a_u\right)$.

At inference step $k$ (corresponding timestep $t_k$), we evaluate the source velocity $v_q = v_u(q_k,t_k\mid I)$ and the expert-guided target velocity with \ac{CFG}
\begin{equation}
    v_p =
    v_u(p_k,t_k\mid I)
    +
    \omega\left[
        v_e(p_k,t_k\mid I)
        -
        v_u(p_k,t_k\mid I)
    \right],
\end{equation}
where \(\omega\) is the \ac{CFG} scale that controls the expert guidance strength. The difference \((v_p-v_q)\) gives the inversion-free editing direction from the agent action chunk toward the expert action chunk.

To encourage source consistency, \flowalign computes one-step clean-action estimates from the source and guided target velocities:
\begin{equation}
    \hat a_u = q_k+(1-t_k)v_q,
    \qquad
    \hat a_e = p_k+(1-t_k)v_p.
\end{equation}
We combine the velocity-difference update with an endpoint-consistency correction that pulls the target estimate toward the source estimate:
\begin{equation}
    a^{(k+1)}
    =
    a^{(k)}
    +
    \eta_{\mathrm{edit}}\,\Delta t_k (v_p-v_q)
    +
    \lambda_{\mathrm{align}}\left(\hat a_u-\hat a_e\right),
\end{equation}
where \(\Delta t_k=t_{k+1}-t_k\), \(\eta_{\mathrm{edit}}\) is the edit scale, and $\lambda_{\mathrm{align}}$ controls the endpoint-consistency correction applied at each iteration.
After \(K\) editing steps, we output $\tilde a = a^{(K)}$. The full algorithm is provided in Appendix Algorithm~\ref{alg:flowalign_editor}. Although \flowalign introduces additional hyperparameters,
the screening sweep shows less than $2$ percentage points of
variation in marginal mean success when the hyperparameters are chosen within the following ranges $\omega\in\{0.2,0.5\}$, $\eta_{\mathrm{edit}}\in\{0.3,0.5,0.7\}$, and $\lambda_{\mathrm{align}}\in\{0.05,0.07,0.10\}$ (Appendix~\ref{app:param_sensitivity}).

\section{Experiments}
\label{sec:experiments}
\label{sec:Experiment}
\begin{tcolorbox}[
    colback=blue!4,
    colframe=blue!45!black,
    boxrule=0.5pt,
    arc=3pt,
    left=8pt, right=8pt,
    top=6pt, bottom=6pt,
    fontupper=\small
]
\textbf{Experimental rationale}

\medskip
\textbf{1. Does \alg improve existing agents?}
Our primary evaluation demonstrates improved task success for
VLA and CaP agents in simulation, and for the VLA on real robots.
These gains come from correcting proposed action chunks while
keeping the underlying agents unchanged.

\medskip
\textbf{2. Why look beyond end-to-end success?}
Many approaches can improve task success, but success alone does
not establish whether OOD action proposals are detected and
corrected before execution, or whether the agent simply recovers
from errors later. Natural agent rollouts lack per-action
ground truth correctness annotations, motivating a controlled perturbation study.

\medskip
\textbf{3. How do we evaluate detection and correction?}
We inject known perturbations into imitation-learning agents'
action chunks. Injection events provide labels for evaluating
detection through precision, recall, and AUROC; task success
with and without correction and OOD gating measures performance
recovery across perturbation types and random seeds.
These labels identify injected changes, which need not always
produce harmful actions.
\end{tcolorbox}

Our primary evaluation asks whether action correction improves the execution of two distinct agent classes---a FLOWER-based \ac{VLA}\citep{reuss2025flower} and a code-as-policy (CaP)~\citep{liang2023code}
agent---without updating the underlying agents. We evaluate agent adaptation on three simulation tasks and further study \ac{VLA} adaptation on two real-robot tasks.

\subsection{Experimental Setup}
\label{sec:experimental-setup}

\paragraph{Tasks and agents.}
We evaluate three simulation tasks: Slalom, CAN, and Keypad, using both FLOWER~\citep{reuss2025flower}-based \ac{VLA} and CaP agents. We additionally evaluate the FLOWER-based \ac{VLA} on two real-robot tasks with a UR5: charger insertion and cup transport around an obstacle (Figure~\ref{fig: paper:exp-tasks-img}).

\paragraph{Expert priors and baselines.}
The expert flows are trained on observation--action pairs \textbf{without explicit goal labels}, and action correction also \textbf{excludes} explicit goal conditioning. The base agent retains its goal or language instruction, while its proposed action chunk supplies the adapter with an intent cue. We compare against unassisted execution, direct sampling from the goal-agnostic expert flow, \textit{To the Noise and Back}~\citep{yoneda2023noise} (DDPM), and \textit{FlashBack CSA}~\citep{sun2025flashback}, adapted to our image-conditioned action-chunk setting.

\begin{figure}[t]
\centering
\begin{subfigure}[t]{0.16\linewidth}
    \centering
    \includegraphics[width=\linewidth,height=3cm,keepaspectratio]{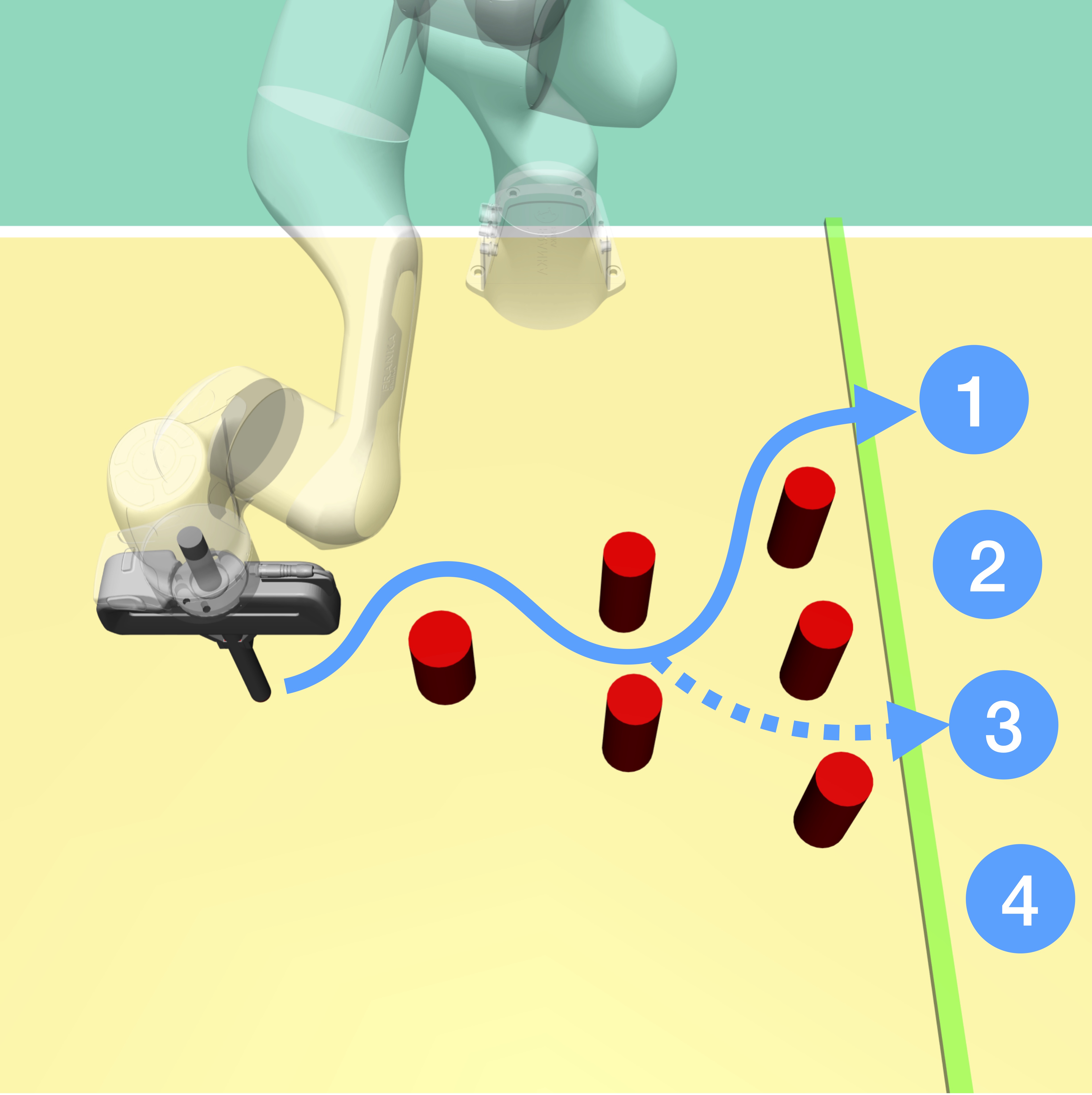}
    \caption{\small Slalom}
    \label{fig:exp-tasks-slalom}
\end{subfigure}
\hfill
\begin{subfigure}[t]{0.16\linewidth}
    \centering
    \includegraphics[width=\linewidth,height=3cm,keepaspectratio]{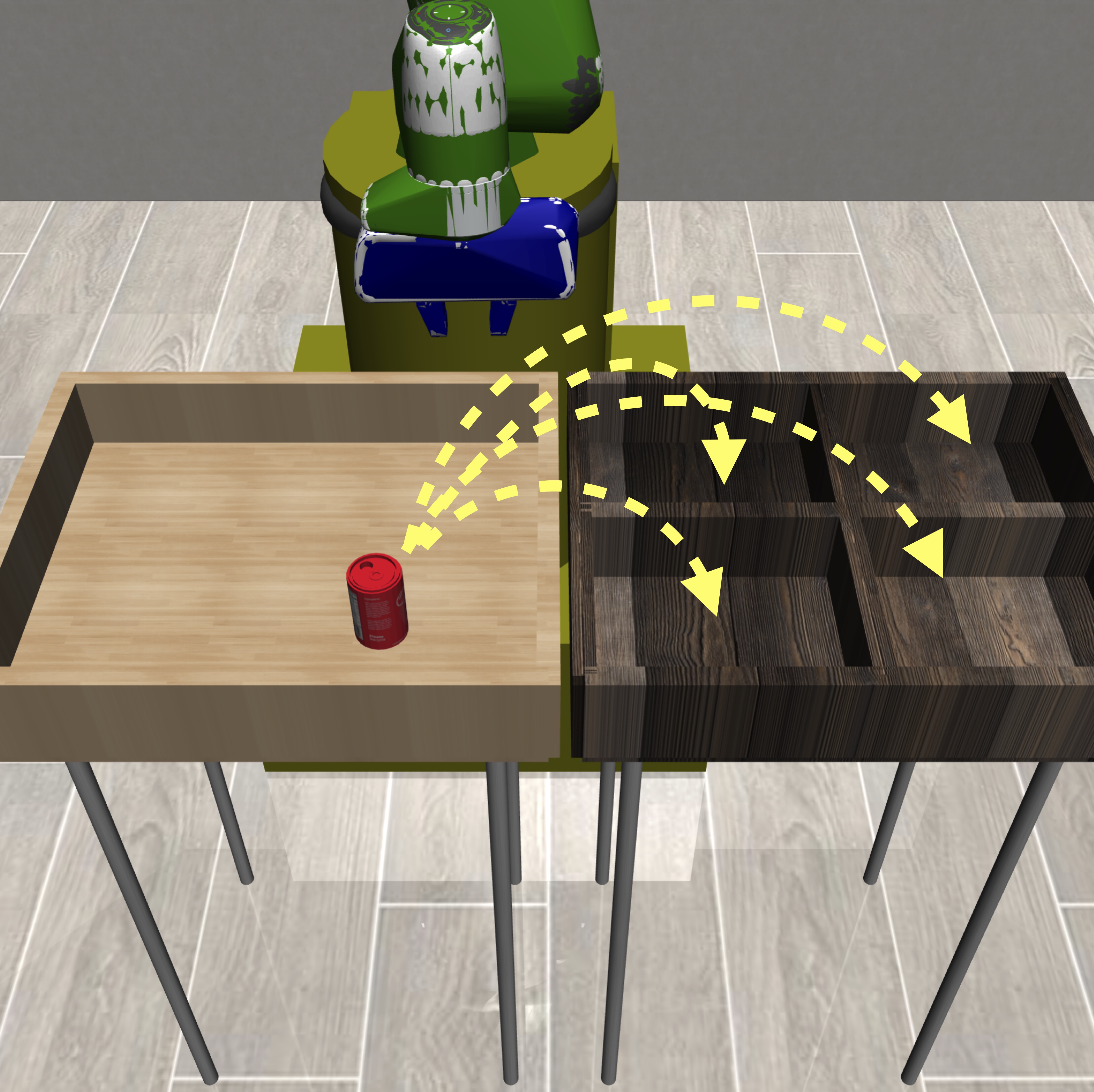}
    \caption{\small CAN}
    \label{fig:exp-tasks-can}
\end{subfigure}
\hfill
\begin{subfigure}[t]{0.16\linewidth}
    \centering
    \includegraphics[width=\linewidth,height=3cm,keepaspectratio]{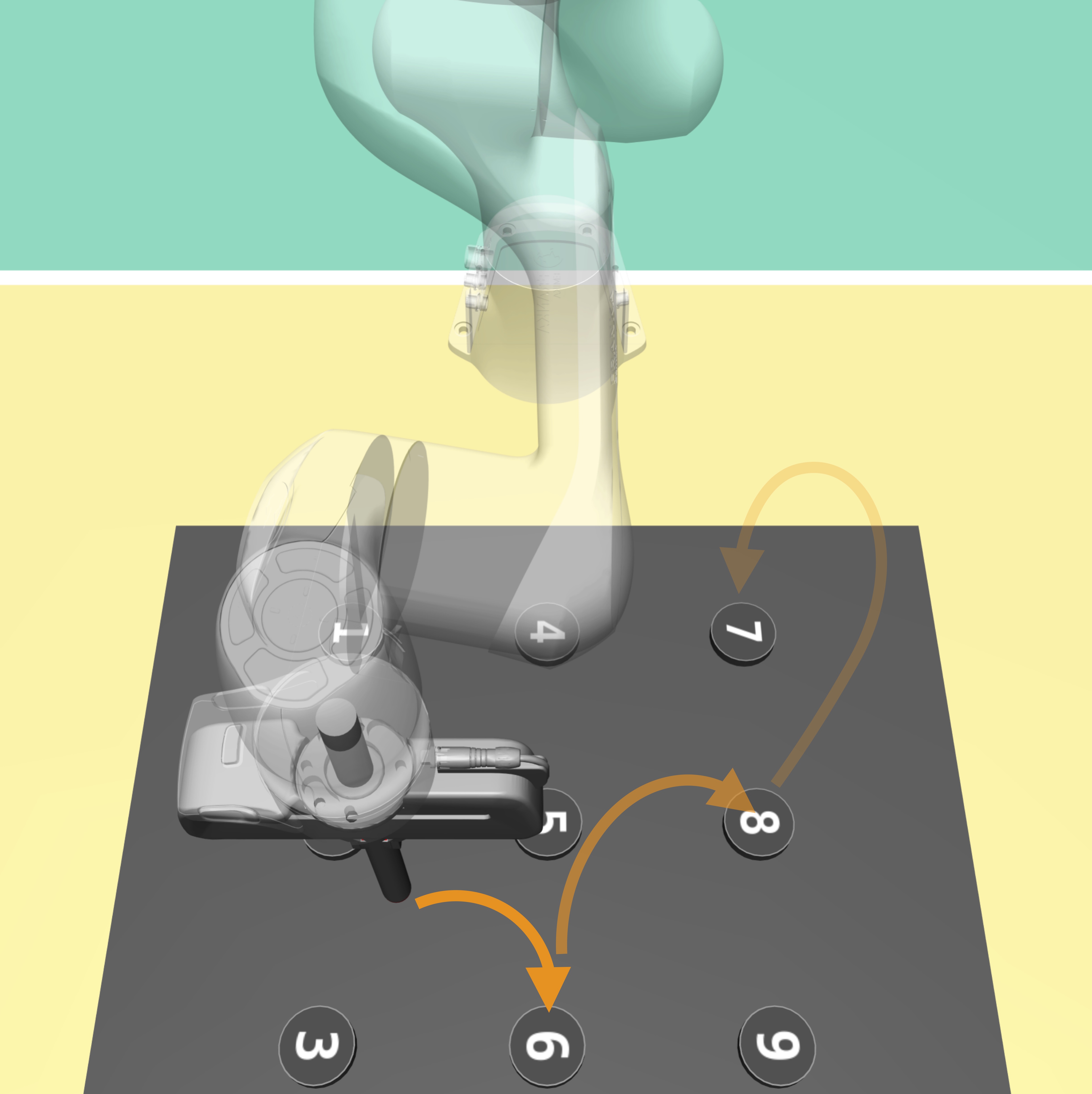}
    \caption{\small Keypad}
    \label{fig:exp-tasks-keypad}
\end{subfigure}
\hfill
\begin{subfigure}[t]{0.16\linewidth}
    \centering
    \includegraphics[width=\linewidth,height=3cm,keepaspectratio]{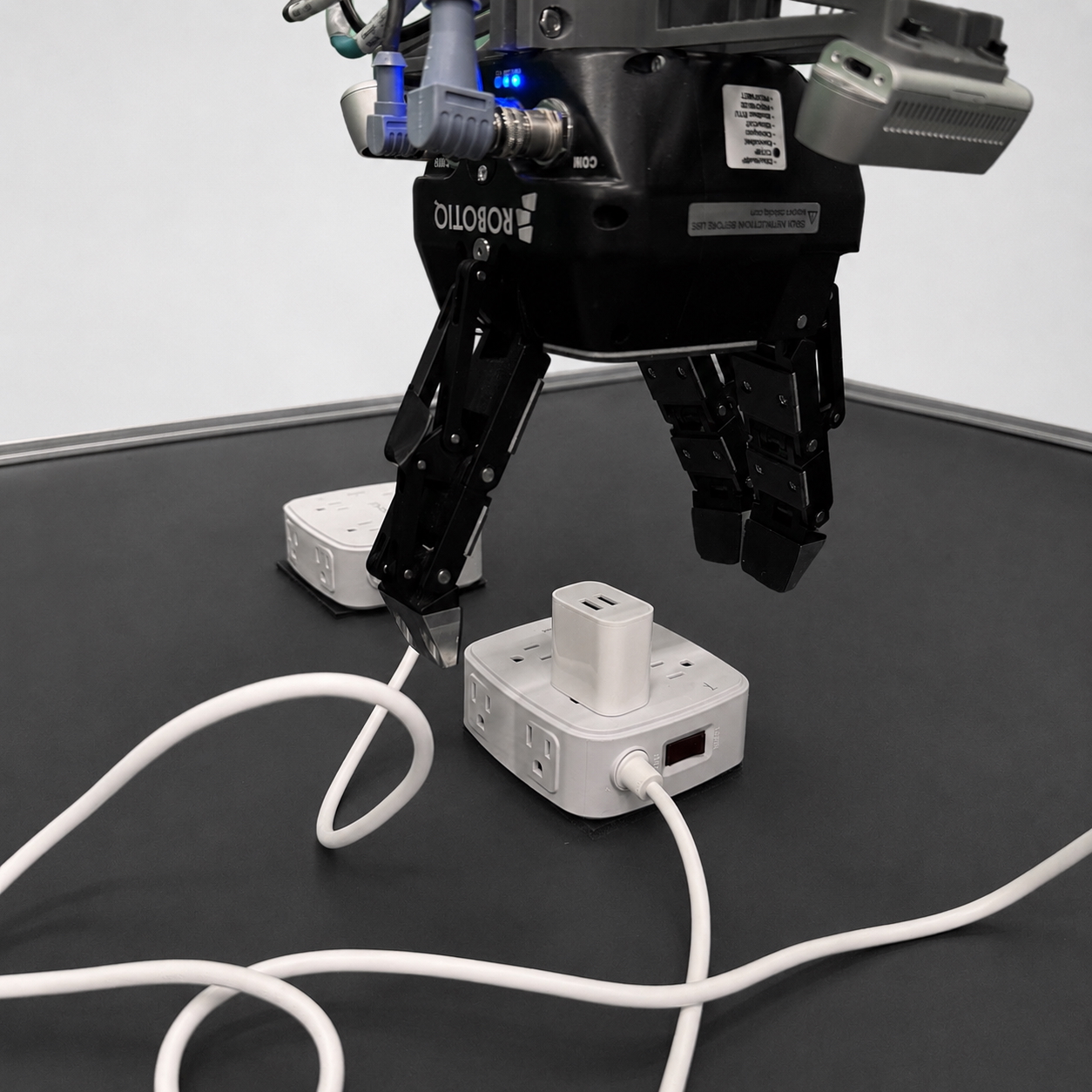}
    \caption{\small Real Charger}
    \label{fig:exp-tasks-charger-real}
\end{subfigure}
\hfill
\begin{subfigure}[t]{0.16\linewidth}
    \centering
    \includegraphics[width=\linewidth,height=3cm,keepaspectratio]{figures/fig9.jpg}
    \caption{\small Serve Cup}
    \label{fig:exp-tasks-cup}
\end{subfigure}
\caption{\small
Experimental tasks. \subref{fig:exp-tasks-slalom} Slalom navigation with $24$ corridor configurations; \subref{fig:exp-tasks-can} CAN placement with four target regions; \subref{fig:exp-tasks-keypad} three-button code entry on a $3 \times 3$ keypad; \subref{fig:exp-tasks-charger-real} real-robot charger insertion; and \subref{fig:exp-tasks-cup} real-robot cup serving with three coaster targets.}
\vspace{-3mm}

\label{fig: paper:exp-tasks-img}
\end{figure}

\subsection{Agent Adaptation in Simulation and on Real Robots}
\label{sec:vla_results}

\paragraph{Simulation with \acs{VLA} and CaP agents.}
Figure~\ref{fig:simulation-agent-stats} reports execution before and after action correction in simulation. On Slalom, \flowalign raises the FLOWER agent's success from $33.06\%$ to $76.67\%$ and the CaP agent's from $63.67\%$ to $97.67\%$. 
For the VLA on Slalom, DDPM, CSA, and \partialflow{}
remain within $2.5$ percentage points of unassisted execution.
Although all three tasks admit multiple goal-dependent behaviors,
Slalom's route choice can be only weakly expressed in a short
local action chunk, making its intent cues vulnerable to partial
noising. In CAN and Keypad, local motion often provides a clearer
cue to the immediate manipulation target or button, even without
revealing the full task goal.
The Slalom results suggest that \flowalign{} better retains these
weak intent cues through source-to-expert transport, without
partially noising the editing state.
On CAN, \partialflow gives the largest gain for CaP ($54.69\%$ to $68.44\%$), while \flowalign with OOD gating gives the largest gain for the \ac{VLA} ($26.25\%$ to $44.70\%$). On Keypad, \partialflow and \flowalign with OOD gating both raise the \ac{VLA} from $46.67\%$ to $62.00\%$; for CaP, \flowalign{} provides the largest gain, reaching
$63.67\%$ with gating compared with $53.67\%$ unassisted. Across the six agent--task pairs, the better \alg editor exceeds every evaluated baseline. These gains are obtained by editing the proposed action chunks while leaving the underlying agent unchanged.

\begin{figure}[t]
    \centering
    \includegraphics[width=0.65\linewidth]{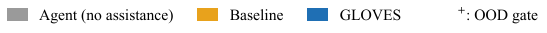}\par\vspace{2pt}
    \begin{minipage}[c]{0.03\linewidth}\centering\rotatebox{90}{\small\bfseries \acs{VLA}}\end{minipage}\hfill
    \begin{subfigure}[c]{0.30\linewidth}
        \centering
        \includegraphics[width=\linewidth]{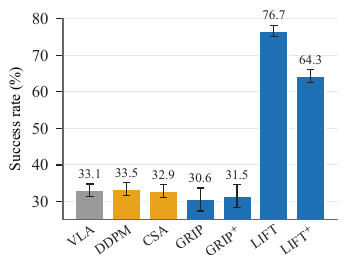}
        \caption{Slalom}\label{fig:simulation-agent-stats-vla-slalom}
    \end{subfigure}
    \hfill
    \begin{subfigure}[c]{0.30\linewidth}
        \centering
        \includegraphics[width=\linewidth]{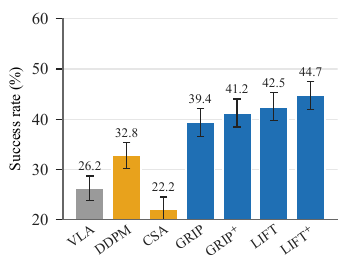}
        \caption{CAN}\label{fig:simulation-agent-stats-vla-can}
    \end{subfigure}
    \hfill
    \begin{subfigure}[c]{0.30\linewidth}
        \centering
        \includegraphics[width=\linewidth]{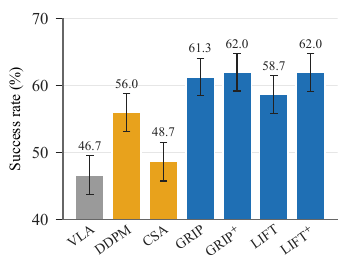}
        \caption{Keypad}\label{fig:simulation-agent-stats-vla-keypad}
    \end{subfigure}

    \vspace{2pt}
    \begin{minipage}[c]{0.03\linewidth}\centering\rotatebox{90}{\small\bfseries CaP}\end{minipage}\hfill
    \begin{subfigure}[c]{0.30\linewidth}
        \centering
        \includegraphics[width=\linewidth]{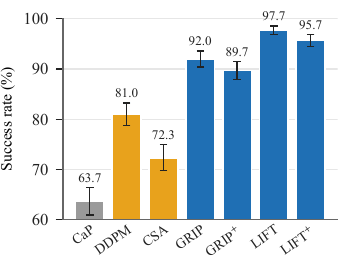}
        \caption{Slalom}\label{fig:simulation-agent-stats-cap-slalom}
    \end{subfigure}
    \hfill
    \begin{subfigure}[c]{0.30\linewidth}
        \centering
        \includegraphics[width=\linewidth]{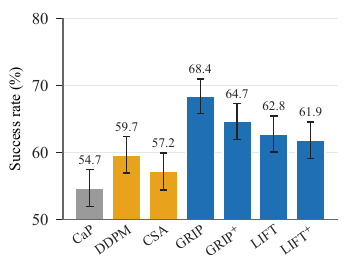}
        \caption{CAN}\label{fig:simulation-agent-stats-cap-can}
    \end{subfigure}
    \hfill
    \begin{subfigure}[c]{0.30\linewidth}
        \centering
        \includegraphics[width=\linewidth]{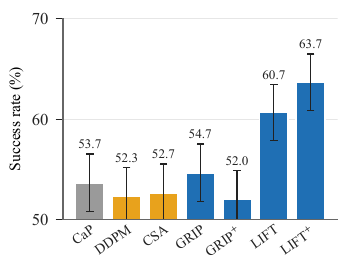}
        \caption{Keypad}\label{fig:simulation-agent-stats-cap-keypad}
    \end{subfigure}
    \caption{
        Simulation success rates (\%) before and after action correction
        for the FLOWER-based \acs{VLA} (top) and the CaP agent (bottom).
        Error bars show the bootstrap standard error over pooled episodes.
        $^{+}$: OOD gating.
        The two agents have different no-assistance baselines, so bars
        should be compared within each panel.
    }
    \vspace{-3mm}
    \label{fig:simulation-agent-stats}
\end{figure}
\begin{wraptable}{r}{0.52\textwidth}
    \vspace{-4mm}
    \centering
    \caption{Real-robot success of the FLOWER-based \acs{VLA} (successful/total trials). Bold: best per task.}
    \label{tab:real-robot-stats}
    \small
    \setlength{\tabcolsep}{3pt}
    \begin{tabular}{lcccc}
        \toprule
        Task & Unassisted & CSA & \acs{GRIP} & \acs{LIFT} \\
        \midrule
        Charger Insertion & $18/40$ & $21/40$ & $\mathbf{40/40}$ & $27/40$ \\
        Cup Serve & $2/30$ & $4/30$ & $\mathbf{22/30}$ & $16/30$ \\
        \bottomrule
    \end{tabular}
    \vspace{-2mm}
\end{wraptable}
\paragraph{\acs{VLA} adaptation on real robots.}
Table~\ref{tab:real-robot-stats} reports the real-world evaluation, which uses the FLOWER-based \ac{VLA}. The two tasks examine complementary scales of correction: charger insertion requires high-precision alignment near contact, whereas cup transport requires sustained corrections to maintain a smooth trajectory around an obstacle and reach the instructed target. On charger insertion, \partialflow improves success from $18/40$ ($45\%$) to $40/40$ ($100\%$); CSA and \flowalign achieve $21/40$ and $27/40$, respectively. On cup transport, successful trials increase from $2$ for the base \ac{VLA} to $22$ with \partialflow and $16$ with \flowalign, compared with $4$ for CSA. Together, these experiments demonstrate precise local correction and sustained trajectory refinement during real-robot execution.

\subsection{What Does the Adapter Correct? A Controlled Perturbation Analysis}
\label{sec:controlled_results}

Section~\ref{sec:vla_results} reports end-to-end success rates, which do not show whether \alg modifies the actions that should be modified. The improvement could also come from recovery after an erroneous action has already been executed. Because agent rollouts do not indicate which actions are erroneous, we construct a controlled setting in which they are known.

\begin{figure}[t]
    \centering

    \begin{minipage}[c]{0.48\linewidth}
    \centering

    \captionof{table}{One-step OOD detection by wrapper.}
    \label{tab:ood_wrapper_main}

    \footnotesize
    \renewcommand{\arraystretch}{1.12}

    \begin{tabular*}{0.96\linewidth}{
        @{\extracolsep{\fill}}lrrr@{}
    }
    \toprule
    Wrapper & Precision & Recall & AUROC \\
    \midrule
    Noised  & .997 & 1.000 & 1.000 \\
    Laggy   & .913 & .344 & .780 \\
    Slow    & .997 & .931 & .988 \\
    Shift   & .994 & .756 & .983 \\
    \midrule
    Macro   & .975 & .758 & .938 \\
    \bottomrule
    \end{tabular*}
    \vspace{3pt}
    \begin{minipage}{0.96\linewidth}
        \centering
        \scriptsize
        Means across three tasks. Midpoint: 1 step, 2 NFE;
        $\alpha=0.005$, macro FPR $=.003$.
        Full metrics and variability:
        Appendix~\ref{app: OOD detection}.
    \end{minipage}
\end{minipage}
    \hfill
    \begin{minipage}[c]{0.50\linewidth}
        \centering
        \includegraphics[width=0.99\linewidth]{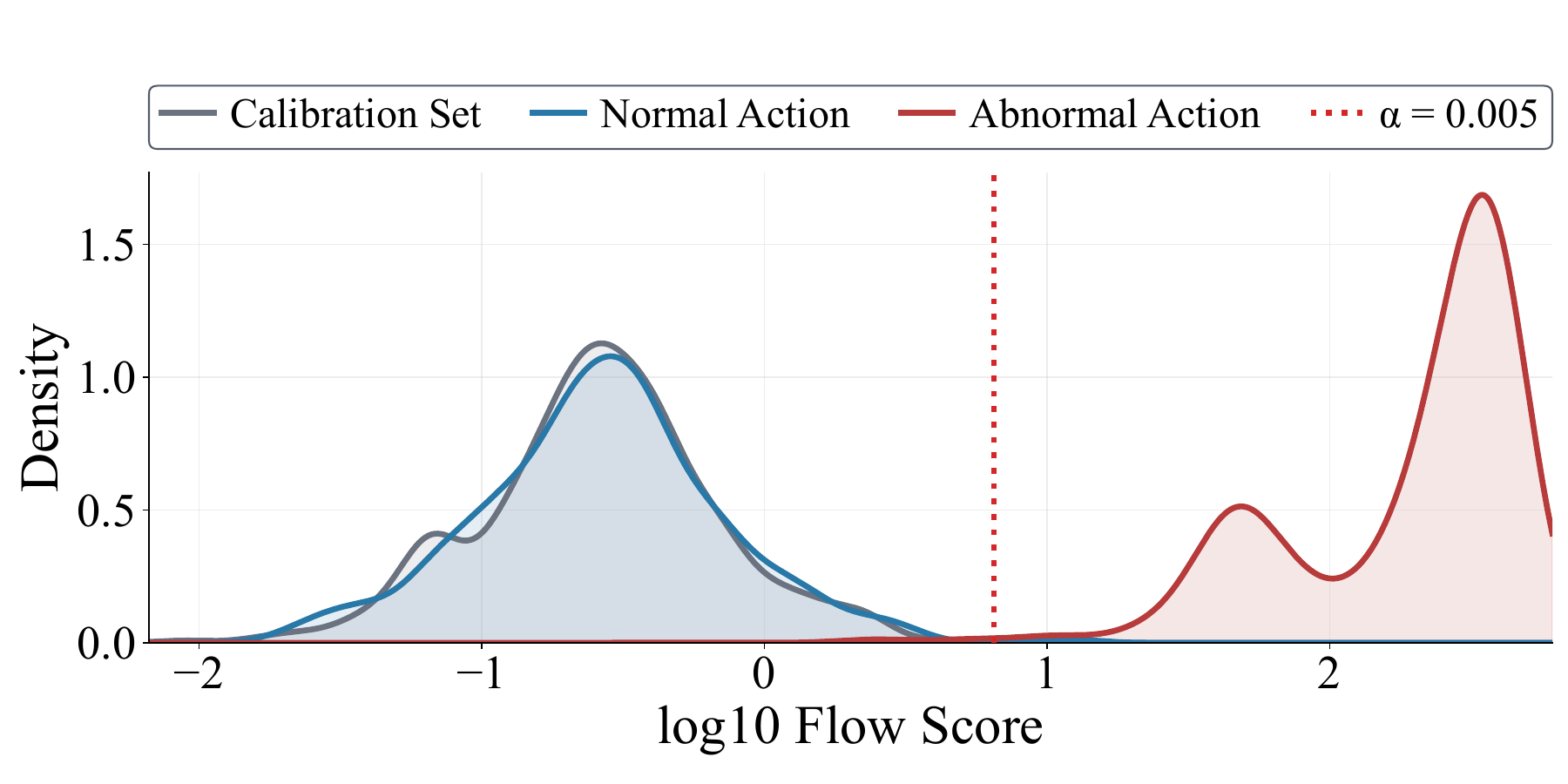}
        \caption{\small
            Reverse-flow score distributions for shift-perturbed Keypad action chunks using one reverse step, evaluated on 250 clean and 250 perturbed chunks.
        }
        \label{fig:one_step_gate_example}
    \end{minipage}
\end{figure}

\begin{table*}[t]
    \centering
    \caption{Task success rates (\%), averaged across four stochastic wrappers.}
    \label{tab:task_success_summary_transposed}
    \begingroup
    \scriptsize
    \setlength{\tabcolsep}{1pt}
    \renewcommand{\arraystretch}{1.28}
    \newcommand{\gaintri}[1]{\textcolor{green!50!black}{\ensuremath{{\scriptscriptstyle\blacktriangle}}\kern.08em{\fontsize{5.2}{6}\selectfont #1}}}
    \newcommand{\losstri}[1]{\textcolor{red!75!black}{\ensuremath{{\scriptscriptstyle\blacktriangledown}}\kern.08em{\fontsize{5.2}{6}\selectfont #1}}}
    \newcommand{\ciann}[1]{\scalebox{0.45}{\ensuremath{\pm\,#1}}}
    \newcommand{\scorecell}[4]{\makebox[6.0em][c]{\makebox[0pt][r]{#1}.\makebox[0pt][l]{#2\ensuremath{{}^{\makebox[0pt][l]{\kern.10em\ciann{#3}}}_{\makebox[0pt][l]{\kern.10em#4}}}}}}
    \begin{tabular*}{\textwidth}{@{\extracolsep{\fill}}l*{4}{c}@{\hspace{8pt}}*{4}{c}@{\hspace{7pt}}}
        \toprule
        Task & Unassisted & Residual & DDPM & CSA & \acs{GRIP} & \acs{GRIP}$^{+}$ & \acs{LIFT} & \acs{LIFT}$^{+}$ \\
        \midrule
        Slalom  & \scorecell{40}{6}{2.25}{} & \scorecell{38}{4}{2.08}{\losstri{2.2}} & \scorecell{24}{8}{1.50}{\losstri{15.8}} & \scorecell{49}{6}{2.42}{\gaintri{9.0}} & \scorecell{41}{9}{2.37}{\gaintri{1.3}} & \scorecell{40}{7}{2.42}{\gaintri{0.1}} & \scorecell{58}{2}{2.42}{\gaintri{17.6}} & \scorecell{\textbf{59}}{\textbf{2}}{2.42}{\gaintri{18.6}} \\
        CAN     & \scorecell{44}{2}{2.58}{} & \scorecell{55}{9}{2.30}{\gaintri{11.7}} & \scorecell{49}{5}{2.19}{\gaintri{5.3}} & \scorecell{58}{8}{2.58}{\gaintri{14.6}} & \scorecell{65}{9}{2.27}{\gaintri{21.7}} & \scorecell{64}{2}{2.27}{\gaintri{20.0}} & \scorecell{62}{9}{2.58}{\gaintri{18.7}} & \scorecell{\textbf{66}}{\textbf{8}}{2.42}{\gaintri{22.6}} \\
        Keypad  & \scorecell{48}{5}{1.00}{} & \scorecell{53}{8}{0.92}{\gaintri{5.3}} & \scorecell{63}{1}{1.08}{\gaintri{14.6}} & \scorecell{62}{4}{0.92}{\gaintri{13.9}} & \scorecell{\textbf{72}}{\textbf{4}}{0.92}{\gaintri{23.9}} & \scorecell{\textbf{72}}{\textbf{4}}{0.87}{\gaintri{23.9}} & \scorecell{67}{2}{1.00}{\gaintri{18.7}} & \scorecell{67}{9}{1.00}{\gaintri{19.4}} \\
\bottomrule
    \end{tabular*}
    \par\vspace{3pt}
    \begin{minipage}{\textwidth}
        \scriptsize
        Equal-weight means across four wrappers. Upper-right $\pm$ values are 95\% bootstrap CI half-widths;
        all columns are recomputed from raw episodes using
        group-stratified resampling within each wrapper ($B=10^4$).
        Lower-right colored triangles
        show changes from Unassisted in percentage points; $+$: OOD gating. Bold: best value per task.
    \end{minipage}
    \endgroup
\end{table*}

\paragraph{Setup.}
To probe the effectiveness of our method, we construct a near-perfect agent by collecting goal-conditioned expert dataset and train a imitation learning (IL) agent (Appendix~\ref{app: IL surrogate}). We further apply one of four stochastic perturbations to the IL agent's proposed action chunks. \textbf{Noised} adds Gaussian noise to every action dimension, \textbf{Laggy} repeats the previous chunk, \textbf{Slow} scales the chunk by a constant factor, and \textbf{Shift} adds a constant offset to one action dimension. Parameters are listed in Appendix Table~\ref{tab:wrapper_params}. The four types of perturbation are widely used in current work~\citep{reddy2018shared,schaff2020residual,yoneda2023noise,mcmahan2024shared,sun2025flashback,kumar2021rma}, which are expected to represent common failure cases in agents. We also compare against a residual correction baseline (Residual), which is given the perturbation types during training. It learns a residual $r_\phi$ that maps a perturbed agent action chunk back to the original one:
\begin{equation}
    \mathcal{L}(\phi)=\mathbb{E}_{(I,a^e)\sim\mathcal{D}_e}\,\mathbb{E}_{c}\,\big\|r_\phi\big(I,c(a^e)\big)-\big(a^e-c(a^e)\big)\big\|_2^2,
    \label{eq:residual_loss}
\end{equation}
where $c$ is the identity or one of the four perturbation types (Appendix~\ref{app:residual_protocol}). At test time, it corrects every proposed chunk as $\tilde a = a_u + r_\phi(I,a_u)$.

This construction lets us separately evaluate whether the gate detects perturbed proposals and whether the editors recover task performance from those perturbations.

\paragraph{One-step OOD scoring.}
For selective intervention, the gate should avoid altering valid actions while adding minimal inference cost, so we prioritize a low false-positive rate and efficient scoring. With one midpoint ODE reverse step, the score achieves a clean false-positive rate of $0.003$ at $\alpha=0.005$, with macro precision $0.975$, recall $0.758$, and AUROC $0.938$ across the three tasks and four perturbation types. Scoring takes $0.356$ ms per sample, compared with $0.625$ ms for five steps, reducing scoring time by approximately $43\%$. We therefore use one step as the default gating setting; Appendix~\ref{app: OOD detection} reports the complete $1/5/50$-step comparison and paired toy visualizations.

The main reduction in wrapper-event recall comes from \textbf{Laggy}: under the balanced evaluation protocol, laggy actions account for approximately $69\%$ of missed perturbation events, with a recall of $0.344$ (Table~\ref{tab:ood_wrapper_main}). This reflects an important distinction between detecting a synthetic perturbation and identifying an action that requires intervention. Because laggy repeats the previous action chunk, a repeated chunk can remain compatible with the current observation despite being labeled as a wrapper event. Wrapper-event recall therefore does not directly measure how often the gate misses actions that would cause execution failure.

Figure~\ref{fig:one_step_gate_example} provides a representative example on Keypad with the \textbf{Shift} perturbation. One-step midpoint scoring separates the perturbed and clean action chunks around the calibrated threshold, rejecting all perturbed chunks while flagging only a small fraction of clean chunks in this example.

\paragraph{Recovery across perturbation types.}
We next ask whether intervention can recover task performance from the controlled action perturbations. Averaged across the four perturbation types, both editors improve over the unassisted agent on every task (Table~\ref{tab:task_success_summary_transposed}), showing that their gains persist after averaging over different failure modes.

We next examine the individual task--wrapper settings to determine whether these average gains are driven by only a subset of perturbations. Across the 12 settings reported here(Appendix ~\ref{app: per wrapper success}), ungated \acs{GRIP} and \acs{LIFT} each outperform the corresponding unassisted agent in 10 and 12 settings, and at least one \alg variant improves performance in all 12. The recovery pattern nevertheless differs across perturbation types. Averaged across tasks, \acs{GRIP} is stronger on Noised and Shift perturbations, whereas \acs{LIFT} is stronger on Laggy and Slow perturbations. The difference is particularly pronounced for Shift, where \acs{GRIP} raises average success from $45.1\%$ to $71.8\%$, while for Laggy, \acs{LIFT} raises it from $46.9\%$ to $57.9\%$. Both editors provide large gains under Slow perturbations.

These results show that the aggregate improvements are not driven by a single perturbation type: recovery is observed across all four controlled failure modes, although the magnitude of the gains varies across editors and perturbations.

\paragraph{Cost of selective intervention.}
Let $r$ denote the intervention rate and $N_{\mathrm{edit}}$ the NFE
of one correction. A one-step midpoint gate costs two evaluations:
\begin{equation}
    N_{\mathrm{gated}}
    = r(2+N_{\mathrm{edit}})+(1-r)2
    = 2+rN_{\mathrm{edit}},
    \qquad
    \Delta N = (1-r)N_{\mathrm{edit}}-2.
\end{equation}
The OOD gate enables selective correction at a measured scoring
cost of $0.356$ ms per sample.
Its net computational benefit depends on the intervention rate
and the cost of action correction.

\subsection{Choosing Between \acs{GRIP} and \acs{LIFT}}
We recommend \partialflow as the default when source-agent rollouts are unavailable or latency and memory are constrained. When agent rollouts and additional compute are available, \flowalign provides an alternative that explicitly models the source action distribution. A small validation set can then determine whether its success gain justifies the additional cost. Appendix~\ref{app:param_sensitivity} reports the parameter sensitivity of both editors.

\vspace{-2mm}
\section{Conclusion}
\label{sec:conclusion}
\vspace{-2mm}

We presented \alg, which improves the actions of a frozen agent without updating it. A goal-agnostic expert flow, learned from limited demonstrations, refines the agent's proposed action chunks, while the proposal itself supplies the task intent. On three simulation tasks, the better \alg editor outperforms every baseline for both a finetuned \ac{VLA} and a code-as-policy agent, and on a real UR5 it raises the \ac{VLA}'s charger-insertion success from $45\%$ to $100\%$. Controlled perturbations show that the flow's reverse score flags perturbed chunks with high precision, and that both editors recover performance across all four perturbation types, beyond a residual corrector trained on those perturbations. \acs{GRIP} needs only the expert flow and is the stronger editor for magnitude and offset errors, whereas \acs{LIFT}, which also models the agent's actions, better preserves intent when the goal-agnostic prior cannot infer it.

\clearpage

\bibliography{references}
\bibliographystyle{preprint}

\clearpage

\appendix

\section{\acl{LIFT} (\acs{LIFT}) Details}
\subsection{\acs{LIFT} Pseudocode}

\begin{algorithm}[h]
\caption{\acs{LIFT}: \acl{LIFT} under the \(x_0\!\to x_1\) Convention}
\label{alg:flowalign_editor}
\small
\begin{algorithmic}[1]
\Require source action \(a_u\), image context \(c\), agent velocity field \(v_u\), expert velocity field \(v_e\)
\Require steps \(K\), time grid \(0=t_0<\cdots<t_K=1\), edit scale \(\eta_{\mathrm{edit}}\), alignment weight \(\lambda_{\mathrm{align}}\), \acs{CFG} scale \(\omega\)
\State \(a^{(0)} \gets a_u\)
\For{\(k=0,\ldots,K-1\)}
    \State \(\Delta t_k \gets t_{k+1}-t_k\), \(\epsilon_k\sim\mathcal{N}(0,I)\)
    \State \(q_k \gets (1-t_k)\epsilon_k+t_k a_u\)
    \State \(p_k \gets q_k+\left(a^{(k)}-a_u\right)\)
    \State \(v_q \gets v_u(q_k,t_k\mid c)\)
    \State \(v_p \gets v_u(p_k,t_k\mid c)+\omega\left[v_e(p_k,t_k\mid c)-v_u(p_k,t_k\mid c)\right]\)
    \State \(\hat a_u \gets q_k+(1-t_k)v_q\), \(\hat a_e \gets p_k+(1-t_k)v_p\)
    \State \(a^{(k+1)} \gets a^{(k)}+\eta_{\mathrm{edit}}\Delta t_k(v_p-v_q)+\lambda_{\mathrm{align}}(\hat a_u-\hat a_e)\)
\EndFor
\State \Return \(\tilde a \gets a^{(K)}\)
\end{algorithmic}
\end{algorithm}

\begin{wrapfigure}{r}{0.50\textwidth}
  \vspace{-\baselineskip} 
  \centering
  \includegraphics[width=0.50\textwidth]{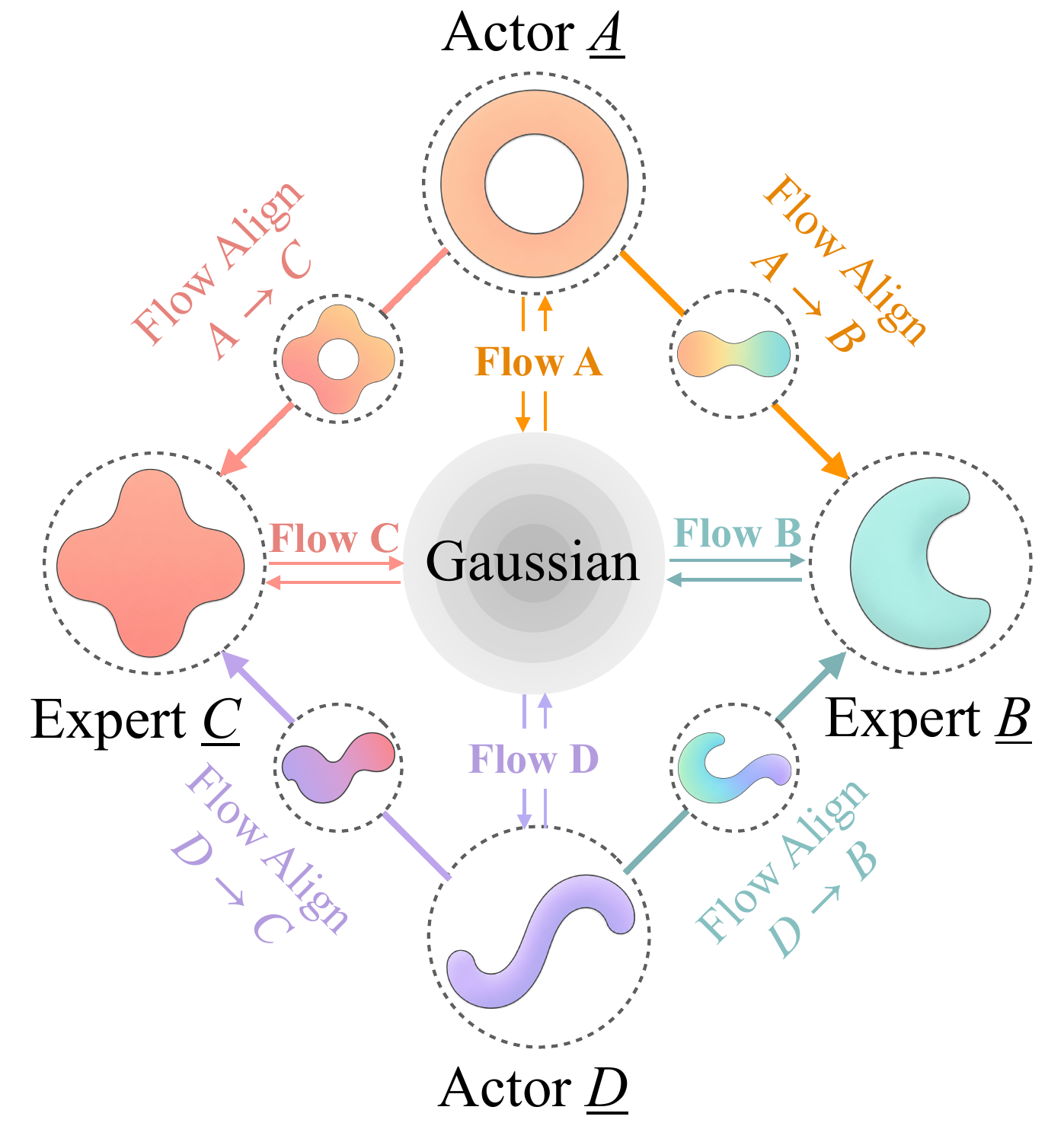}
  \caption{\alg provides transformations between imperfect agent and  expert distributions.}
  \vspace{5mm}
  \label{fig:conceptual-flow-align}
\end{wrapfigure}

\subsection{Convention translation.}
Our implementation uses the \(x_0\!\to x_1\) convention, where \(x_0=\epsilon\sim\mathcal N(0,I)\) is noise and \(x_1=a\) is the clean action chunk:
\[
    x_t=(1-t)\epsilon+t a,
    \qquad
    v(x_t,t)=a-\epsilon .
\]
Some FlowAlign derivations use the opposite convention, where the clean sample is indexed as \(x_0\) and noise as \(x_1\). This is only a reparameterization of the same path. If \(s=1-t\), then
\[
    x_t=(1-t)\epsilon+t a
    =
    (1-s)a+s\epsilon,
    \quad
    v_t=-v_s .
\]
Therefore, the two conventions are equivalent as long as the velocity sign, endpoint estimate, and update step are translated consistently. In our convention, the clean endpoint estimate is
\[
    \hat a = x_t+(1-t)v(x_t,t),
\]
and the FlowAlign update uses a positive \(\Delta t_k=t_{k+1}-t_k\).
If the implementation uses \(\sigma=1-t\), then \(\Delta t_k=-\Delta\sigma_k\).

\section{Experimental Setup}
\label{app:detail exp setup}

\subsection{Simulation Environments}
\label{app:sim}

We evaluate \alg on three simulated robot-assistance benchmarks that jointly cover low-dimensional planar navigation, contact-rich manipulation and long-horizon sequential interaction. The three tasks were chosen so that the action-adaptation problem is exercised across qualitatively different action spaces (from $2$-D Cartesian increments to $8$-D joint-delta-plus-gripper commands), visual modalities (single- versus dual-view RGB, with or without proprioceptive state), and mode structures (small discrete sets of corridors and placement regions, large combinatorial sequence spaces, and continuously parameterized goal poses). All environments follow a Gym-style interface with a fixed per-episode budget; an episode terminates upon task completion, upon a task-specific failure condition (collision, wrong-button activation, or out-of-tolerance contact, as applicable), or upon reaching the maximum number of environment steps. Unless stated otherwise, RGB observations are rendered at a resolution of $256 \times 256$, action chunks have horizon $T = 10$, and an emitted chunk is unrolled open-loop in the environment before the next policy query is issued. Table~\ref{tab:sim-tasks} summarizes the key dimensions of each task; the remainder of this section provides per-task descriptions of the environment, action interface, observation interface, mode/goal structure, success and failure criteria, and expert demonstration set.

\begin{table}[h]
\centering
\small
\setlength{\tabcolsep}{4pt}
\begin{tabular}{@{}p{0.13\linewidth}p{0.14\linewidth}p{0.20\linewidth}p{0.13\linewidth}p{0.15\linewidth}p{0.06\linewidth}p{0.06\linewidth}@{}}
\hline
\textbf{Task} & \textbf{Simulator} & \textbf{Action space} & \textbf{Visual input} & \textbf{Modes / goals} & \textbf{Horiz.} & \textbf{Demos} \\
\hline
Slalom & MuJoCo & $2$-D $\Delta xy$ & $1$ side camera & $24$ corridor layouts & $300$ & $400$ \\
CAN & Robosuite  & $7$-D delta EEF + gripper & side and wrist camera & $4$ placement regions & $400$ & $400$ \\
Keypad & MuJoCo & $3$-D $\Delta xyz$ & side and wrist camera & $9^3 = 729$ button codes & $1000$ & $450$ \\
\hline
\end{tabular}
\caption{Summary of the three simulation benchmarks used to evaluate \alg.}
\label{tab:sim-tasks}
\end{table}

\subsection{Slalom}
\label{app:avoiding}

We adopt the obstacle-avoidance environment from the D3IL benchmark, implemented in MuJoCo, in which a planar manipulator must traverse a corridor populated by six pillars arranged between a start line and a goal line. The robot controls planar end-effector displacements via $2$-D $\Delta xy$ actions with fixed height and orientation. The environment exposes $24 = 2 \times 3 \times 4$ obstacle-layout modes. These modes act as a discrete categorical label that determines which corridor must be traversed for the episode to count as a success. Observations consist of a single third-person RGB image at $256 \times 256$ resolution. An episode terminates when the end-effector crosses the goal line in the prescribed corridor (success), when contact with any pillar is detected (collision), or when the per-episode budget of $300$ environment steps is exhausted (timeout). The expert dataset contains $400$ trajectories ($100$ per goal-region label), recorded as observation-action pairs with per-trajectory mode annotations.

For evaluation, we run 30 episodes for each random seed, and each model is evaluated over 10 random seeds.

\subsection{RoboMimic CAN}
\label{app:can}

We evaluate the can placement task from RoboMimic, executed in the Robosuite simulator on a MuJoCo backend. A Franka Emika Panda arm must transport a soup can from a randomly initialized position into one of four discrete target placement regions. The four regions $\{\text{top-left, top-right, bottom-left, bottom-right}\}$ are exposed as distinct task modes, providing a multimodal expert distribution rather than the single-mode protocol of the original benchmark. Actions are $7$-dimensional delta end-effector pose, and a $1$-D continuous gripper command (open at $-1$, closed at $+1$). Observations consist of two RGB streams: an external third-person camera and a wrist-mounted camera, both rendered at $256 \times 256$ resolution and updated at the $20\,\mathrm{Hz}$ control frequency. An episode terminates when the simulator reports a successful placement of the can within the requested region (success) or after $400$ environment steps (timeout). The expert dataset comprises $400$ trajectories ($100$ per region, totalling $60{,}399$ transitions) collected from a region-conditioned scripted planner with mild target-pose noise ($\sigma_x = \sigma_y = 0.05\,\mathrm{m}$) to induce diversity around each placement target.

For evaluation, we run 32 episodes for each random seed, and each model is evaluated over 10 random seeds.

\subsection{Keypad}
\label{app:keypad}

We introduce a custom MuJoCo manipulation task in which the robot must enter a $3$-button code on a $3 \times 3$ keypad in a prescribed order. Actions are $3$-D Cartesian displacements; the low-level controller appends a constant downward-facing quaternion so that the end-effector consistently approaches the keypad face from above. The task admits $9^3 = 729$ distinct modes, each corresponding to a three-digit button sequence drawn uniformly per episode. Observations include both a keypad-facing RGB camera and a wrist-mounted RGB camera, each rendered at $256 \times 256$. An episode terminates when the requested sequence is successfully entered (success), when an incorrect button is depressed at any sub-step (wrong-button failure), upon undesired contact with the keypad frame (collision), or after $1000$ environment steps (timeout). The expert dataset contains $450$ demonstrations ($50$ per single-button press across $9$ buttons), each averaging approximately $66$ control steps; multi-press episodes are obtained at training time by composing single-press demonstrations.

A distinguishing property of this task is that the language condition is \emph{per-timestep} rather than per-trajectory: the ``current target button'' advances each time a sub-goal is completed within the same episode. The policy is therefore supplied with a one-hot encoding of the \emph{current} target button only, while the image-conditioned assistive policy receives no language signal and must infer the active sub-goal from RGB observations alone.

For evaluation, we run 30 episodes for each random seed, and each model is evaluated over 10 random seeds.

\subsection{Agent Policy}
\subsubsection{Imitation Learning Policy with wrapper}
\label{app: IL surrogate}
The first type of agent policy used in simulation is a goal-conditioned imitation learning policy. It shares the same architecture as the flow matching model, with the only difference being that the IL agent additionally takes the goal as input. We then apply stochastic perturbation wrappers to the agent policy in order to inject structured failure modes into its actions.

Formally, each wrapper transforms an action chunk
$
a \in \mathbb{R}^{B \times T \times A},
$
where $B$ is the number of parallel environments, $T=10$ is the chunk horizon, and $A$ is the action dimension. Randomness is sampled independently for each environment using decorrelated child generators derived from a shared \texttt{numpy.random.Generator}.

\vspace{0.5em}
\noindent\textbf{Noised Wrapper.}
The noised wrapper injects i.i.d. Gaussian perturbations into each action dimension:
$$
a_i \leftarrow a_i + \epsilon,
\quad
\epsilon \sim \mathcal{N}(0, \sigma^2 I).
$$
This models imprecise or jittery control behavior.

\vspace{0.5em}
\noindent\textbf{Laggy Wrapper.}
With probability $p_{\mathrm{rep}}$, the laggy wrapper replays the previous action chunk instead of emitting a newly predicted chunk. This simulates delayed reactions, stalls, or persistent human corrections.

\vspace{0.5em}
\noindent\textbf{Slow Wrapper.}
With probability $p_{\text{slow}}$, the entire action chunk is scaled by a factor $\alpha < 1$:
$$
a_i \leftarrow \alpha a_i.
$$
This models hesitant or low-confidence behavior where the intended motion direction is correct but under-actuated.

\vspace{0.5em}
\noindent\textbf{Shift Wrapper.}
With probability $p_{\text{shift}}$, a constant offset $\alpha$ is added to one action dimension $d$ across the full chunk. This introduces systematic control bias, such as calibration errors or directional drift.

\begin{table}[h]
\label{wrapper-table}
\centering
\caption{Stochastic-wrapper parameter configurations used for the three simulation
  tasks. Each cell specifies the parameter tuple for the
  named wrapper.
}
\label{tab:wrapper_params}
\small
\setlength{\tabcolsep}{8pt}
\renewcommand{\arraystretch}{1.15}
\begin{tabular}{lcccc}
\toprule
& \textbf{Noised} & \textbf{Laggy} & \textbf{Slow} & \textbf{Shift} \\
\textbf{Task} & ($\sigma$) & ($p_{\mathrm{rep}}$) & ($\alpha,\,p_{\text{slow}}$) & ($d,\,\alpha,\,p_{\text{shift}}$) \\
\midrule
Slalom       & $0.005$ & $0.20$ & $(0.5,\,0.5)$ & $(1,\,0.005,\,0.5)$ \\
Keypad         & $0.015$ & $0.30$ & $(2.0,\,0.5)$ & $(1,\,0.005,\,0.5)$ \\
RoboMimic CAN  & $0.15$  & $0.20$ & $(0.5,\,1.0)$ & $(1,\,0.20,\,0.5)$  \\
\bottomrule
\end{tabular}
\end{table}

\subsubsection{Vision Language Action Model}
\label{app:vla-finetuning}

We finetune FLOWER for the CAN, Slalom, Charger Real, and Cup Real \ac{VLA}-agent experiments using task-specific demonstrations. These runs use a latent-space Gaussian-blur curriculum with a linear schedule, following FACTR~\citep{liu2025factr}. To strengthen instruction following at inference, we apply language dropout during finetuning and combine the language-conditioned and language-dropped velocity predictions using \acl{CFG}~\citep{ho2022classifierfree,nichol2021glide}:
\[
v_{\mathrm{cfg}}(x_t,t,c,L)
=
v_\theta(x_t,t\mid c,\varnothing)
+
s_{\mathrm{cfg}}
\left[
v_\theta(x_t,t\mid c, L)
-
v_\theta(x_t,t\mid c,\varnothing)
\right],
\]
where \(c\) denotes the visual observation, \(L\) the language instruction, \(\varnothing\) the language-dropped condition, and \(s_{\mathrm{cfg}}\) the guidance scale. The finetuning settings and available \ac{CFG} inference settings are summarized in Table~\ref{tab:vla-finetuning}.

\begin{table}[h]
\centering
\scriptsize
\setlength{\tabcolsep}{2pt}
\begin{tabularx}{\linewidth}{@{}lXXXX@{}}
\toprule
\textbf{Setting} & \textbf{CAN} & \textbf{Slalom} & \textbf{Charger Real} & \textbf{Cup Real} \\
\midrule
Learning rate & \(1\times 10^{-5}\) & \(1\times 10^{-5}\) & \(1\times 10^{-5}\) & \(1\times 10^{-5}\) \\
Batch / GPUs & \(24\times2\times4=192\) & \(40\times1\times4=160\) & \(24\times2\times4=192\) & \(24\times2\times4=192\) \\
Training length & 50 epochs & 50 epochs & 50 epochs & 50 epochs\\
Latent curriculum & \multicolumn{4}{c}{Gaussian blur, linear schedule} \\
Language dropout & \multicolumn{4}{c}{\(0.2\)} \\
\acs{CFG} scale \(s_{\mathrm{cfg}}\) & \(1.5\) & \(2.0\) & 1.0 & 2.0 \\
\bottomrule
\end{tabularx}
\caption{FLOWER finetuning settings and \acs{CFG} inference hyperparameters for the \acs{VLA} agent experiments.}
\label{tab:vla-finetuning}
\end{table} 
The exact training-time language instructions are:
\begin{itemize}
    \item \textbf{CAN:} ``Pick up the can and place it in the top left of the destination bin.''; ``Pick up the can and place it in the top right of the destination bin.''; ``Pick up the can and place it in the bottom left of the destination bin.''; ``Pick up the can and place it in the bottom right of the destination bin.''
    \item \textbf{Slalom:} We use three space-separated mode integers as the instruction. The mode set is \(\{0,1\}\times\{0,1,2\}\times\{0,1,2,3\}\), rendered as strings such as \texttt{0 0 0} through \texttt{1 2 3}.
    \item \textbf{Charger Real:} ``Insert the charger plug into the left target socket.''; ``Insert the charger plug into the right target socket.''
    \item \textbf{Cup Real:} ``Put the cup on the top coaster.''; ``Put the cup on the middle coaster.''; ``Put the cup on the bottom coaster.''
\end{itemize}

\subsubsection{Code-as-Policy}
\label{app:cap-construction}

We construct demonstration-guided code-as-policy (CaP) agents for CAN, Slalom, and Keypad, inspired by Code as Policies~\citep{liang2023code}. A language-model coding agent develops task-specific Python controllers that receive goal instructions and predict action chunks from image observations. The CaP can develop and iterate with demonstration data and rollout feedback. This extends the idea of expressing robot policies as executable programs to our demonstration-guided setting. The evaluated CaPs and fitted parameters are frozen; execution requires no language-model calls or online code generation.

Each CaP uses ten demonstration trajectories containing RGB images, actions and corresponding language instructions. Generally, the coding agent would fit local mappings from actions to image-feature displacements using regularized least squares, and extract task-specific waypoints, interaction offsets, or motion profiles. Visual servoing uses these mappings to correct image-space errors. We allow at most ten rounds of rollout feedback for CaP to optimize per task. No neural action policy is trained. Table~\ref{tab:cap-construction} summarizes the common settings.

At runtime, CaP uses the frozen SAM ViT-B segmentation, prompted by deterministic image-based heuristics and combined with geometric image processing, to extract object and tool locations. CAN and Keypad additionally use a wrist view; Keypad uses detected button radii as a visual cue for approach and retraction. CaP maintains internal phase and execution history.

\begin{table}[h]
    \centering
    \scriptsize
    \setlength{\tabcolsep}{2pt}
    \begin{tabularx}{\linewidth}{@{}lXXX@{}}
    \toprule
    \textbf{Setting} & \textbf{CAN} & \textbf{Slalom} & \textbf{Keypad} \\
    \midrule
    Demonstrations & 10 & 10 & 10 \\
    
    RGB views used & Main + wrist & Main & Main + wrist \\
    Image processing size & \multicolumn{3}{c}{\(256\times256\) per view} \\
    Action dimension & 7 & 2 & 3 \\
    Prediction / execution & \multicolumn{3}{c}{\(H=10\), \(K=10\)} \\
    Runtime language-model calls & \multicolumn{3}{c}{None} \\
    \bottomrule
    \end{tabularx}
    \caption{Construction and execution settings for the CaP agents. Action
    dimensions follow each environment's existing control interface.}
    \label{tab:cap-construction}
\end{table}

The task-specific CaPs operate as follows:
\begin{itemize}
    \item \textbf{CAN:} The instruction specifies one of four destination-bin
    regions. Main- and wrist-view features guide a phase-based controller
    through approach, grasp, lift, transport, descent, and release.
    Demonstrations provide contact and transport offsets, while visual
    feedback governs alignment and phase progression.
    \item \textbf{Slalom:} The instruction is a route tuple in
    \(\{0,1\}\times\{0,1,2\}\times\{0,1,2,3\}\).
    Image-detected obstacles define route-dependent waypoints. A fitted
    visual servo generates planar actions, and observed tool position
    determines progress through the route.
    \item \textbf{Keypad:} The instruction specifies an ordered button
    sequence. A demonstration-derived grid and visual servo guide the tool
    to each button. Demonstration-derived vertical press and retract
    profiles are combined with visual contact and clearance checks.
    Phase bookkeeping is synchronized with the actions actually issued,
    including any truncated final chunk, before advancing to the next
    button.
\end{itemize}

If required visual features are unavailable, the CaPs attempt
reacquisition and issue bounded recovery chunks, preserving the commanded
gripper state where applicable. Persistent loss of required visual evidence
beyond the 30-step recovery allowance terminates the rollout as a perception
failure.

\subsection{Residual Correction Baseline}
\label{app:residual_protocol}
No paired agent and expert actions are available, so the residual network $r_\phi$ in Equation~\ref{eq:residual_loss} is trained on perturbations of the expert's own chunks $a\in\mathbb{R}^{T\times d_a}$. For each training sample, one operator $c$ is drawn: the identity with probability $0.25$, which gives a zero target, and otherwise one of four perturbation families chosen uniformly at random. The families are never composed:
\begin{equation*}
\begin{aligned}
c_{\mathrm{noise}}(a)&=a+\varepsilon, & \varepsilon&\sim\mathcal{N}(0,\sigma^{2}I), & \sigma&\sim\mathrm{LogU}[0.001,\,1.0],\\
c_{\mathrm{scale}}(a)&=\alpha\,a, & & & \alpha&\sim\mathrm{LogU}[0.01,\,2.0],\\
c_{\mathrm{shift}}(a)&=a+s\,\beta\,\mathbf{1}_T e_d^{\top}, & d&\sim\mathcal{U}\{1,\dots,d_a\},\ s\sim\mathcal{U}\{\pm1\}, & \beta&\sim\mathrm{LogU}[0.01,\,2.0],\\
c_{\mathrm{lag}}(a)_t&=a_{t-m\Delta}, & m&\sim\mathcal{U}\{1,\dots,5\}, & &
\end{aligned}
\end{equation*}
where $\mathrm{LogU}$ denotes a log-uniform distribution and $\Delta$ is the number of steps executed per chunk; the lagged chunk is taken from the same demonstration. These ranges are identical for all tasks and are not set to the evaluation parameters in Table~\ref{tab:wrapper_params}.

The residual network has the same architecture and parameter count as the task's expert flow, with a constant time input so that $r_\phi$ is a deterministic map. We train it with batch size $256$, learning rate $3\times10^{-4}$, weight decay $10^{-3}$, and $50$ epochs, hold out $10\%$ of the trajectories for validation, and use the best validation checkpoint. At test time, the corrected chunk $a_u+r_\phi(I,a_u)$ is clipped to the action bounds and executed at every query, without OOD gating. Evaluation uses the same perturbations, seeds, and episode counts as the other methods.

\subsection{Real World Experiment}

\begin{table}[h]
\centering
\small
\setlength{\tabcolsep}{4pt}
\begin{tabular}{@{}p{0.14\linewidth}p{0.22\linewidth}p{0.22\linewidth}p{0.11\linewidth}p{0.07\linewidth}p{0.10\linewidth}@{}}
\hline
\textbf{Task} & \textbf{Action space} & \textbf{Visual input} & \textbf{Targets} & \textbf{Demos} & \textbf{Transitions} \\
\hline
UR5 Charger & $3$-D Cartesian setpoint & side + wrist camera & $2$ sockets & $200$ & $22{,}797$ \\
UR5 Cup     & $7$-D end-effector pose & side + wrist camera & $3$ cups & $235$ & $45{,}409$ \\
\hline
\end{tabular}
\caption{Summary of the two real-world UR5 tasks used to evaluate \alg. }
\label{tab:ur5-data}
\end{table}

\paragraph{UR5 Charger Insertion}

The robot must align and insert a charger into one of two sockets under visual feedback from one side camera and one wrist camera. This task stresses fine pose correction near contact, where small translation errors translate to missed insertions. The dataset contains 200 successful scripted trajectories (100 per socket) recorded at $10\,\text{Hz}$, totaling $22{,}797$ control steps. 

Evaluation criterion. A trial is considered successful if the agent inserts the charger into the target socket specified by the language instruction within 160 environment steps. Failures include unsuccessful insertion, misalignment, or excessive contact forces that trigger an emergency stop. At the beginning of each episode, the charger’s pose is reset by grasping it from a fixed initial location. For each experimental setting, we evaluate 40 episodes in total (20 episodes per target socket) and report the success rate.

\paragraph{UR5 Fluid Carrying}

The task requires the robot to transport a cup to one of three target locations while avoiding a fixed obstacle. The dataset contains 235 successful demonstrations, approximately balanced across the three targets, totaling 45,409 control steps.

Evaluation criterion. A trial is successful if the agent places the cup at the target pose specified by the language instruction within 160 environment steps. Failure cases include a positional error greater than 0.025 m in the x or y direction, a positional error greater than 0.015 m in the z direction, or any collision with the obstacle.

\section{Model Architecture}
\label{app:arch}

We instantiate the assistive policy with a \ac{DiT} velocity-field architecture (Sec.~\ref{app:dit}) coupled to a shared visual and state encoder stack (Sec.~\ref{app:enc}). Table~\ref{tab:hp-summary} summarizes the architecture hyperparameters and training details.

\subsection{\acs{DiT} Velocity Field}
\label{app:dit}

The \ac{DiT} velocity field is a Transformer that operates on a token set composed of (i) the chunked action tokens of the noisy sample $x_t$ and (ii) a sequence of context tokens generated by the visual and state encoder stack. Each block consists of three pre-LayerNorm sub-layers: self-attention over action tokens with rotary positional embeddings applied to queries and keys; cross-attention from action queries to context keys and values without positional encoding on the context; and a GELU-activated MLP with expansion ratio $r \in \{2, 4\}$.

Each sub-layer is conditioned via adaptive layer normalization with a zero-initialized residual gate (AdaLN-Zero): the time embedding is mapped by a zero-initialized linear projection to per-layer scale and gate parameters. The time embedding itself is sinusoidal with frequency dimension $256$ and is fed through a two-layer MLP to the hidden dimension; we scale $t$ by $1000$ before the sinusoidal embedding, in line with prevailing diffusion- and flow-matching practice. 

\subsection{Visual and State Encoders}
\label{app:enc}

We adopt an ImageNet-pretrained ResNet-$18$ with frozen batch normalization as the visual backbone in all main experiments. Features are extracted from the third residual stage (channel dimension $256$, spatial stride $16$); for a $256 \times 256$ input this yields a $16 \times 16$ token grid, i.e., $256$ spatial tokens per view. A $1 \times 1$ convolution then projects the feature channels to the context dimension $d_c$. The backbone is fine-tuned end-to-end.

When multiple views are present, we instantiate \emph{independent (non-weight-shared) encoders} per view, and the per-view token grids are concatenated along the sequence axis before being passed to the \ac{DiT} cross-attention.

The proprioceptive state encoder is a three-layer feedforward network from input dimension $S$ to width $256$ to width $256$ to output dimension $d_c$, and it produces a single context token consumed alongside the visual tokens.

\subsection{Training Procedure}
\label{app:train}

Table~\ref{tab:hp-summary} consolidates the model and training hyperparameters used across all reported experiments. Both the standard flow-matching model and the FlowAlign model  share the optimizer, and regularization settings in (b); their architectural sizes are listed separately in (a).

\begin{table}[h]
    \centering
    \small
    \setlength{\tabcolsep}{3pt}
    \begin{minipage}[t]{0.48\linewidth}
        \centering
        \textbf{(a) Model hyperparameters} \\[0.4em]
        \begin{tabular}{@{}lr@{}}
        \hline
        \textbf{Component} & \textbf{Default} \\
        \hline
        Visual backbone                       & ResNet-$18$ \\
        Pretrained weights                    & ImageNet \\
        Feature stage                         & 3rd residual \\
        Feature channels                      & $256$ \\
        Spatial tokens per view               & $16 \times 16$ \\
        Image normalization                   & ImageNet stats \\
        State encoder hidden width            & $256$ \\
        FM context dim $d_c$                  & $256$ \\
        FM hidden width $d$                   & $128$ \\
        FM MLP ratio $r$                      & $4$ \\
        FlowAlign context dim $d_c$            & $512$ \\
        FlowAlign hidden width $d$             & $512$ \\
        FlowAlign MLP ratio $r$                & $2$ \\
        Attention heads $H$                   & $4$ \\
        Block dropout                         & $0.1$ \\
        Time embedding dim                    & $256$ \\
        Time scaling factor                   & $1000$ \\
        Visual dropout $p_{\mathrm{vd}}$      & $0.1$ \\
        Action chunk length $T$               & $10$ \\
        \hline
        \end{tabular}
    \end{minipage}
    \hfill
    \begin{minipage}[t]{0.48\linewidth}
        \centering
        \textbf{(b) Training hyperparameters} \\[0.4em]
        \begin{tabular}{@{}lr@{}}
        \hline
        \textbf{Component} & \textbf{Default} \\
        \hline
        Optimizer                             & AdamW \\
        Base learning rate                    & $3 \times 10^{-4}$ \\
        Visual learning rate                  & $3 \times 10^{-5}$ \\
        Visual LR scale                       & $0.1\times$ \\
        Weight decay                          & $10^{-3}$ \\
        Adam $(\beta_1, \beta_2)$             & $(0.9,\ 0.999)$ \\
        Batch size                            & $256$ \\
        Training epochs                       & $500$ \\
        \acs{CFG} dropout (training) $p_{\mathrm{cfg}}$ & $0.2$ \\
        Time-sampling schedule                & Uniform \\
        Karras $p_{\mathrm{std}}$             & $1.2$ \\
        \hline
        \end{tabular}
    \end{minipage}
    \caption{Default model (a) and training (b) hyperparameters used across all tasks. }
    \label{tab:hp-summary}
\end{table}

\subsection{Parameter Sensitivity}
\label{app:param_sensitivity}

\paragraph{\acs{LIFT}.}
The screening sweep shows limited marginal sensitivity within evaluated ranges of all three editing parameters.
Mean success rates vary by $1.10$, $1.98$, and $0.17$ percentage points over $\omega\in\{0.2,0.5\}$,
$\eta_{\mathrm{edit}}\in\{0.3,0.5,0.7\}$, and
$\lambda_{\mathrm{align}}\in\{0.05,0.07,0.10\}$, respectively, averaging over task--wrapper settings and remaining parameters.

\paragraph{\acs{GRIP}.}
\acs{GRIP} fixes $\lambda_{\text{align}}=1$ and selects $t_{\mathrm{init}}$ for each setting. Table~\ref{tab:energy_ablation} isolates the energy term by comparing $\lambda_{\text{align}}=0$ with $\lambda_{\text{align}}=1$ at the same $t_{\mathrm{init}}$. The energy term improves success in $9$ of the $12$ settings; in the remaining three it changes success by at most $2.0$ percentage points. The largest gain is on Keypad with the Noised perturbation ($+20.7$).

\begin{table}[!ht]
\centering
\caption{Ablation of the guidance strength for \acs{GRIP}.
}
\label{tab:energy_ablation}
\small
\setlength{\tabcolsep}{5pt}
\renewcommand{\arraystretch}{1.1}
\begin{tabular}{llcccc}
\toprule
\textbf{Task} & \textbf{Wrapper} & \textbf{Baseline} & $\lambda=0$ & $\lambda=1$ & $\Delta$ \\
\midrule
\multirow{4}{*}{Slalom}
  & noised  & $40.0 \pm 2.9$ & $45.3 \pm 2.9$ & $46.7 \pm 5.8$ & $+1.3$            \\
  & laggy   & $38.7 \pm 2.8$ & $33.7 \pm 2.7$ & $31.7 \pm 3.8$ & $-2.0$            \\
  & slow    & $55.7 \pm 2.9$ & $49.3 \pm 2.9$ & $49.3 \pm 5.6$ & $\phantom{+}0.0$  \\
  & shift   & $28.0 \pm 2.6$ & $34.3 \pm 2.8$ & $40.0 \pm 3.5$ & $+5.7$       \\
\cmidrule(lr){1-6}
\multirow{4}{*}{Keypad}
  & noised  & $34.7 \pm 2.7$ & $30.7 \pm 2.7$ & $51.3 \pm 7.8$ & $+20.7$      \\
  & laggy   & $54.0 \pm 2.9$ & $57.7 \pm 2.9$ & $61.3 \pm 6.3$ & $+3.7$       \\
  & slow    & $52.7 \pm 2.9$ & $93.3 \pm 1.4$ & $95.0 \pm 2.6$ & $+1.7$            \\
  & shift   & $52.7 \pm 2.9$ & $81.0 \pm 2.3$ & $82.0 \pm 3.5$ & $+1.0$            \\
\cmidrule(lr){1-6}
\multirow{4}{*}{CAN}
  & noised  & $50.9 \pm 2.8$ & $52.8 \pm 2.8$ & $52.5 \pm 5.7$ & $-0.3$            \\
  & laggy   & $48.1 \pm 2.8$ & $51.2 \pm 2.8$ & $55.0 \pm 6.3$ & $+3.8$       \\
  & slow    & $23.1 \pm 2.3$ & $58.1 \pm 2.8$ & $62.8 \pm 3.5$ & $+4.7$       \\
  & shift   & $54.7 \pm 2.8$ & $92.8 \pm 1.4$ & $93.4 \pm 3.3$ & $+0.6$            \\
\midrule
\multicolumn{2}{l}{\textit{Mean across 12 cells}}
              & $42.9$         & $55.5$         & $58.9$         & $+3.5$            \\
\bottomrule
\end{tabular}
\end{table}

\subsection{Per-Wrapper Success Rates}
\label{app: per wrapper success}
\begin{table*}[h]
    \centering
    \caption{\alg success rates (\%) on stochastic-wrapper agents compared with DDPM and CSA. The best and second-best results are highlighted in \besthl{blue} and \secondhl{green}, respectively.}
    \label{tab:flowalign_best_triplet_comparison}
    \small
    \setlength{\tabcolsep}{3.5pt}
    \resizebox{\textwidth}{!}{\begin{tabular}{ll|c|cc|cccc}
    \toprule
    \multirow{2}{*}{Task} & \multirow{2}{*}{Agent} & \multirow{2}{*}{\shortstack{Agent\\Performance}} & \multirow{2}{*}{DDPM} & \multirow{2}{*}{CSA} & \multicolumn{2}{c}{GRIP} & \multicolumn{2}{c}{LIFT} \\
    \cmidrule(lr){6-7}\cmidrule(lr){8-9}
    & & & & & w/o OOD & w/ OOD & w/o OOD & w/ OOD \\
    \midrule
    \multirow{4}{*}{Slalom}
      & laggy  & $38.7{\pm}5.8$ & $15.3{\pm}2.2$ & \besthl{$43.6{\pm}5.0$} & $31.7{\pm}3.8$ & $31.0{\pm}3.7$ & \secondhl{$39.4{\pm}5.4$} & $39.3{\pm}5.3$ \\
      & noised & $40.0{\pm}7.1$ & $33.3{\pm}3.3$ & $42.7{\pm}5.3$ & $46.7{\pm}5.8$ & $45.7{\pm}5.8$ & \besthl{$53.7{\pm}5.3$} & \secondhl{$53.6{\pm}5.4$} \\
      & slow   & $55.7{\pm}5.1$ & $35.7{\pm}3.3$ & $42.3{\pm}4.0$ & $49.3{\pm}5.6$ & $49.3{\pm}5.6$ & \besthl{$74.6{\pm}4.6$} & \secondhl{$74.0{\pm}4.4$} \\
      & shift  & $28.0{\pm}2.8$ & $15.0{\pm}2.3$ & \secondhl{$69.7{\pm}4.0$} & $40.0{\pm}3.5$ & $36.7{\pm}4.0$ & $65.0{\pm}4.7$ & \besthl{$70.0{\pm}4.7$} \\
    \midrule
    \multirow{4}{*}{CAN}
      & laggy  & $48.1{\pm}8.2$ & $53.8{\pm}5.3$ & $51.5{\pm}5.3$ & $55.0{\pm}6.3$ & $53.1{\pm}6.5$ & \secondhl{$66.6{\pm}5.1$} & \besthl{$81.9{\pm}4.4$} \\
      & noised & $50.9{\pm}5.6$ & $1.6{\pm}1.4$ & \besthl{$60.3{\pm}5.0$} & $52.5{\pm}5.7$ & $50.6{\pm}5.8$ & \secondhl{$55.9{\pm}5.2$} & $55.6{\pm}5.0$ \\
      & slow   & $23.1{\pm}6.0$ & $62.2{\pm}5.0$ & $70.9{\pm}4.8$ & $62.8{\pm}3.5$ & $60.9{\pm}4.8$ & \besthl{$72.2{\pm}4.7$} & \secondhl{$71.6{\pm}4.7$} \\
      & shift  & $54.7{\pm}3.9$ & $80.3{\pm}4.2$ & $52.5{\pm}5.5$ & \besthl{$93.4{\pm}3.3$} & \secondhl{$92.2{\pm}3.2$} & $56.9{\pm}5.4$ & $58.1{\pm}5.6$ \\
    \midrule
    \multirow{4}{*}{Keypad}
      & laggy  & $54.0{\pm}8.1$ & $57.3{\pm}2.3$ & $64.0{\pm}2.3$ & $61.3{\pm}6.3$ & $62.3{\pm}6.6$ & \besthl{$82.0{\pm}1.7$} & \secondhl{$81.7{\pm}1.7$} \\
      & noised & $34.7{\pm}5.6$ & $47.7{\pm}2.3$ & $25.7{\pm}1.7$ & \besthl{$51.3{\pm}7.8$} & \secondhl{$50.7{\pm}6.9$} & $46.0{\pm}2.7$ & $44.8{\pm}2.4$ \\
      & slow   & $52.7{\pm}6.2$ & $81.3{\pm}2.0$ & $92.0{\pm}1.0$ & \secondhl{$95.0{\pm}2.6$} & \besthl{$96.0{\pm}2.1$} & $79.3{\pm}2.0$ & $88.0{\pm}2.0$ \\
      & shift  & $52.7{\pm}5.6$ & $66.0{\pm}2.0$ & $68.0{\pm}2.0$ & \besthl{$82.0{\pm}3.5$} & \secondhl{$80.7{\pm}4.0$} & $61.4{\pm}1.7$ & $57.3{\pm}2.0$ \\
    \bottomrule
    \end{tabular}}
\end{table*}

\subsection{Number of Expert Demonstrations}
\label{app:demo_ablation}
We retrain the expert flow of \acs{GRIP} on nested subsets of $20$, $50$, $100$, and $200$ demonstrations drawn from the same dataset. All other settings follow the full-data model: the validation split and normalization statistics are shared, each model trains for $500$ epochs, and we use the best validation checkpoint. Evaluation reuses the settings of the reported runs, including $t_{\mathrm{init}}$, so only the flow checkpoint changes. Table~\ref{tab:demo_ablation} reports success averaged over the four perturbation types. With $50$ demonstrations, \acs{GRIP} reaches about $60\%$ of its full-data gain over the unassisted agent, and with $200$ demonstrations about $80\%$. CAN depends most on the amount of data. On Slalom, where the full-data gain is within noise, success shows no consistent trend with the number of demonstrations.

\begin{table}[!ht]
\centering
\small
\caption{Success rate (\%) of \acs{GRIP} with expert flows trained on different numbers of demonstrations, averaged over the four perturbation types. Full is the reported model trained on all demonstrations (Table~\ref{tab:sim-tasks}).}
\label{tab:demo_ablation}
\begin{tabular}{lcccccc}
\toprule
Task & Unassisted & 20 & 50 & 100 & 200 & Full \\
\midrule
Slalom & 40.6 & 43.3 & 42.5 & 40.8 & 43.6 & 41.9 \\
CAN    & 44.2 & 45.8 & 53.4 & 54.3 & 59.1 & 65.9 \\
Keypad & 48.5 & 60.1 & 65.0 & 65.1 & 69.0 & 72.4 \\
\midrule
Average & 44.4 & 49.7 & 53.6 & 53.4 & 57.2 & 60.1 \\
\bottomrule
\end{tabular}
\end{table}

\section{Out of distribution detection and toy example}
\label{app: OOD detection}
\paragraph{Conditional Reverse-Flow Score}
We evaluate whether the conditional reverse-flow score used by our OOD detector produces meaningful conformal acceptance regions in controlled two-dimensional settings. Let \(v_\theta(x_t,t\mid c)\) be a conditional rectified flow trained to transport samples from a standard Gaussian prior \(\mathcal{P}_Z=\mathcal{N}(0,I)\) to the class-conditional data distribution \(\mathcal{P}_X(x\mid c)\). Given a test point \(x\) and condition \(c\), 
we run the learned flow backward for $N$ steps to obtain an approximate reversed latent $\hat z_N$ and non-conformity score $s_N$
\begin{equation*}
    \hat z_N(x;c)
    =\operatorname{ODE}^{-1}_{N}\!\left[v_\theta,x,1\!\rightarrow\!0;c\right],
    \qquad
    s_N(x;c)=\|\hat z_N(x;c)\|_2^2,
    \label{eq:app_midpoint_score}
\end{equation*}

where $N$ denotes the number of reverse integration steps. The robotic experiments use the image context $I$; the toy examples use a class condition $c$. For each integration budget, calibration and test examples are scored with the same numerical procedure. Given $n$ calibration scores $s_{N,i}$, we compute

\begin{equation*}
    p_N(x;c)
    =\frac{1+\sum_{i=1}^{n}\mathbf{1}\{s_{N,i}\geq s_N(x;c)\}}{n+1}.
    \label{eq:app_midpoint_calibration}
\end{equation*}
Low scores indicate that the point maps to a typical region of the Gaussian prior under the condition-dependent reverse map, while high scores indicate a conditionally unlikely sample. The gate flags a proposal when $p_N\leq\alpha$. Calibration scores are
computed before inference; only the test score and its rank are needed
online. We use one midpoint step as the efficiency-oriented default and
compare it with five and fifty steps below.

\subsection{Detection on Controlled Action Perturbations}

\begin{wrapfigure}{r}{0.4\textwidth}
  \centering
  \includegraphics[width=0.4\textwidth, height=4cm, keepaspectratio]{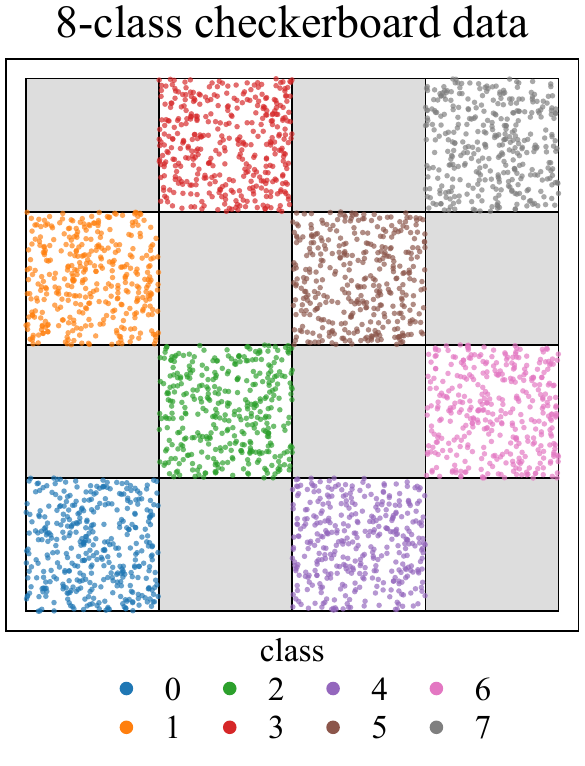}
  \caption{Eight-class checkerboard distribution used for conditional OOD evaluation.}
  \label{fig:appendix-multiclass-training}
\end{wrapfigure}

We evaluate the four implemented wrappers (\textit{noised},
\textit{laggy}, \textit{slow}, and \textit{shift}) on Slalom, CAN,
and Keypad with 3 different reverse steps 1/5/50. Each task--wrapper--step setting contains 250 clean and 250 perturbed test chunks. Each task--step setting uses 500 expert calibration examples, with $\alpha=0.005$. The evaluation uses the trained
\ac{DiT} with EMA weights. Reported aggregate metrics are equal-weight means over the 12 task--wrapper settings; a positive label records a wrapper activation, rather than a verified action failure.

\begin{table}[t]
    \centering
    \small
    \setlength{\tabcolsep}{3.5pt}
    \caption{Reverse-step sensitivity with midpoint integration. Metrics
    are macro-averaged over the 12 task--wrapper settings. Time is the
    benchmarked scoring time per sample, excluding action editing.}
    \label{tab:app_reverse_steps}
    \begin{tabular}{@{}rrrrrrrrr@{}}
    \toprule
    Steps & NFE & Precision & Recall & F1 & FPR & AUROC & AUPRC & Time (ms) \\
    \midrule
    1  & 2   & 0.975 & 0.758 & 0.805 & 0.003 & 0.938 & 0.948 & 0.356 \\
    5  & 10  & 0.989 & 0.847 & 0.884 & 0.005 & 0.957 & 0.966 & 0.625 \\
    50 & 100 & 0.989 & 0.818 & 0.852 & 0.001 & 0.954 & 0.962 & 3.359 \\
    \bottomrule
    \end{tabular}
\end{table}

One-step scoring achieves high precision and a low clean false-positive
rate while reducing scoring time by approximately $43\%$ relative to
five steps (Table~\ref{tab:app_reverse_steps}). Five steps improve
wrapper-event recall and ranking metrics; increasing the budget to
fifty steps leaves macro AUROC at $0.954$ while increasing scoring time
to $3.359$ ms. These measurements characterize the cost and detection
behavior of each numerical score.

\begin{table}[t]
    \centering
    \small
    \setlength{\tabcolsep}{4pt}
    \caption{One-step midpoint detection by wrapper. Each entry is the
    mean $\pm$ standard deviation across the three tasks.}
    \label{tab:app_ood_by_wrapper}
    \begin{tabular}{@{}lccccc@{}}
    \toprule
    Wrapper & Precision & Recall & F1 & AUROC & AUPRC \\
    \midrule
    Noised  & $0.997\pm0.002$ & $1.000\pm0.000$ & $0.999\pm0.001$ & $1.000\pm0.000$ & $1.000\pm0.000$ \\
    Laggy   & $0.913\pm0.142$ & $0.344\pm0.305$ & $0.454\pm0.383$ & $0.780\pm0.042$ & $0.820\pm0.068$ \\
    Slow    & $0.997\pm0.003$ & $0.931\pm0.120$ & $0.960\pm0.068$ & $0.988\pm0.021$ & $0.992\pm0.014$ \\
    Shift   & $0.994\pm0.007$ & $0.756\pm0.416$ & $0.808\pm0.327$ & $0.983\pm0.030$ & $0.980\pm0.034$ \\
    \bottomrule
    \end{tabular}
\end{table}

Laggy accounts for approximately $66\%$ of missed wrapper events under
one-step scoring (Table~\ref{tab:app_ood_by_wrapper}). This wrapper
repeats a preceding expert action chunk, which can remain compatible
with the current observation during continuous motion. \textbf{Consequently,
wrapper-event recall is not the fraction of unsafe actions detected.}

The lower laggy recall is consistent with this distinction, although
these labels do not establish whether individual missed chunks are
harmless. Downstream rollout success provides a separate assessment of
the combined gate and editor.

\paragraph{Toy Distributions}

We consider an 8-class checkerboard distribution on \([-2,2]^2\). The \(4\times 4\) grid contains data only in alternating cells, yielding eight occupied cells and eight empty cells (Figure~\ref{fig:appendix-multiclass-training}).

Each occupied cell is assigned one class label, and the conditional flow receives the normalized class label \(c=k/7\). This distribution is useful because a correct conditional detector should accept points inside the queried class cell and reject both empty cells and points from other class cells when they are evaluated under the wrong condition.

\paragraph{Results}

The accepted-region plots (Figure~\ref{fig:appendix-8class-acceptance}) show that the learned reverse-flow score produces class-specific conformal acceptance regions. For each query label \(c=k/7\), the accepted region \(p(x,c)>\alpha\) concentrates in the corresponding occupied checkerboard cell, while empty cells and other class cells are rejected. This supports the use of the conditional reverse-flow prior score as a lightweight OOD detector.

\begin{figure}[t]
    \centering
    {\small\textbf{One step}\par}
    \includegraphics[width=\linewidth]{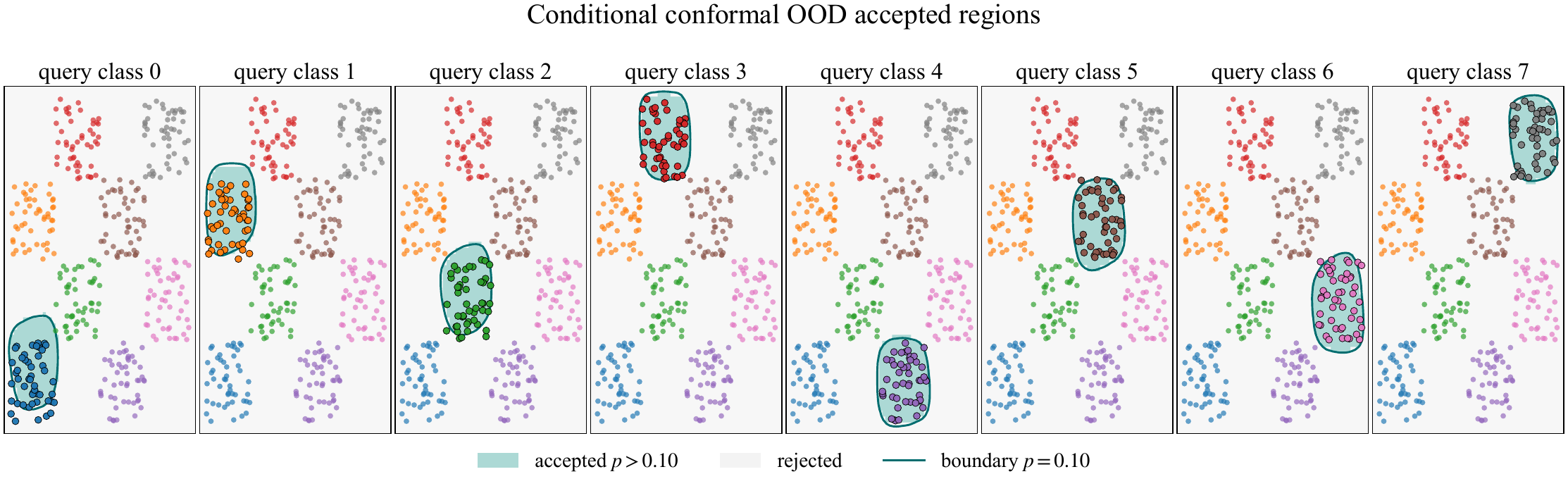}
    \medskip
    {\small\textbf{Five steps}\par}
    \includegraphics[width=\linewidth]{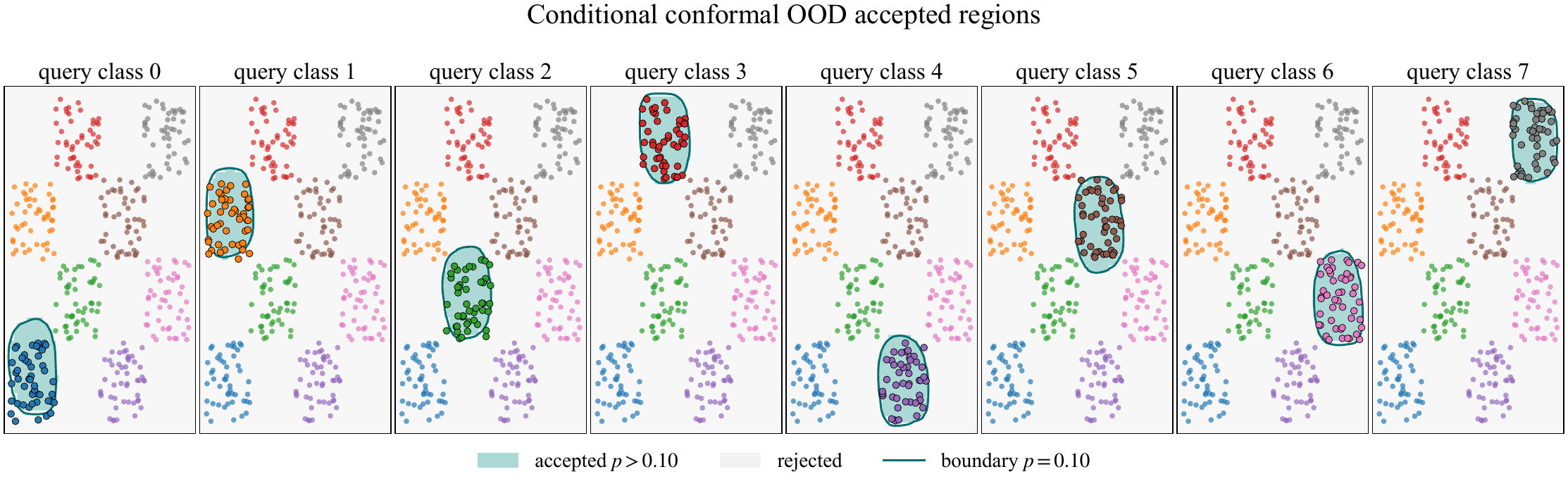}
    \caption{Conditional acceptance regions at $\alpha=0.10$ using one
    reverse step (top) and five steps (bottom). Shaded regions
    satisfy $p_N(x;c)>\alpha$; contours mark the acceptance boundary.
    Each panel uses the displayed class as the query condition.}
    \label{fig:appendix-8class-acceptance}
\end{figure}

We next test partial calibration coverage by fixing the query label to class 7 and progressively removing classes from the calibration set (Figure~\ref{fig:appendix-ID-hist}). When class 7 is included in calibration, the acceptance rate inside the class-7 cell is high. When class 7 is absent, performance persists, because the conditional reverse-flow scores remain comparable across classes.

\begin{figure}[t]
    \centering
    {\small\textbf{One step}\par}
    \includegraphics[width=\linewidth]{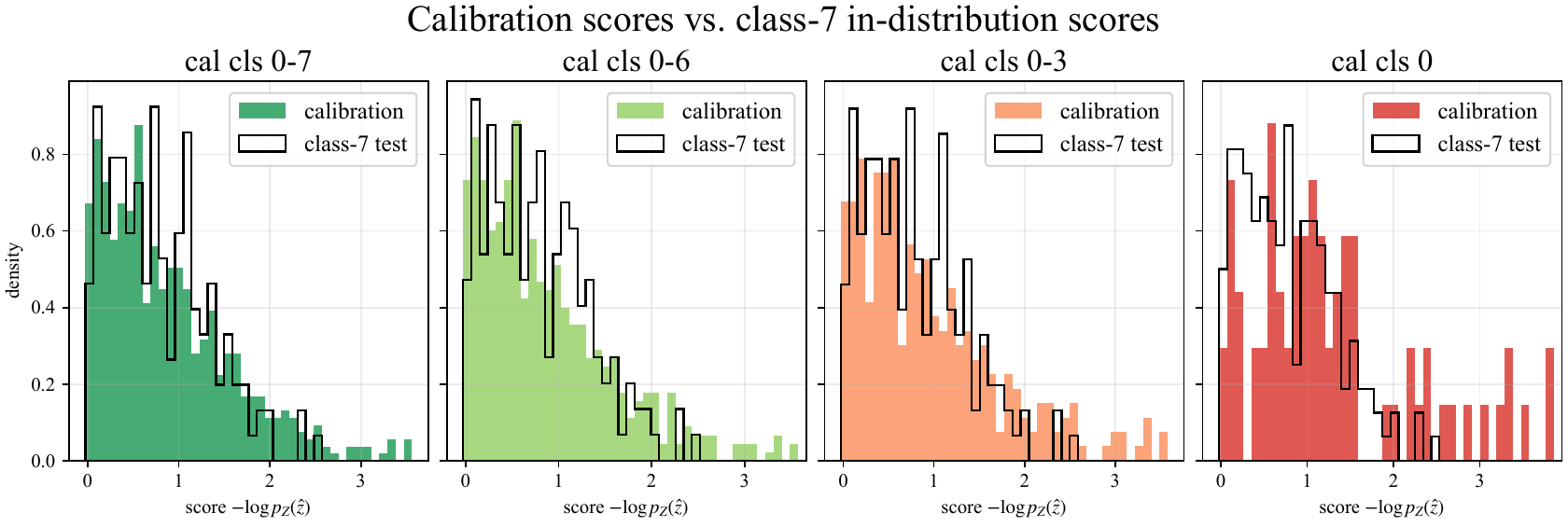}
    \medskip
    {\small\textbf{Five steps}\par}
    \includegraphics[width=\linewidth]{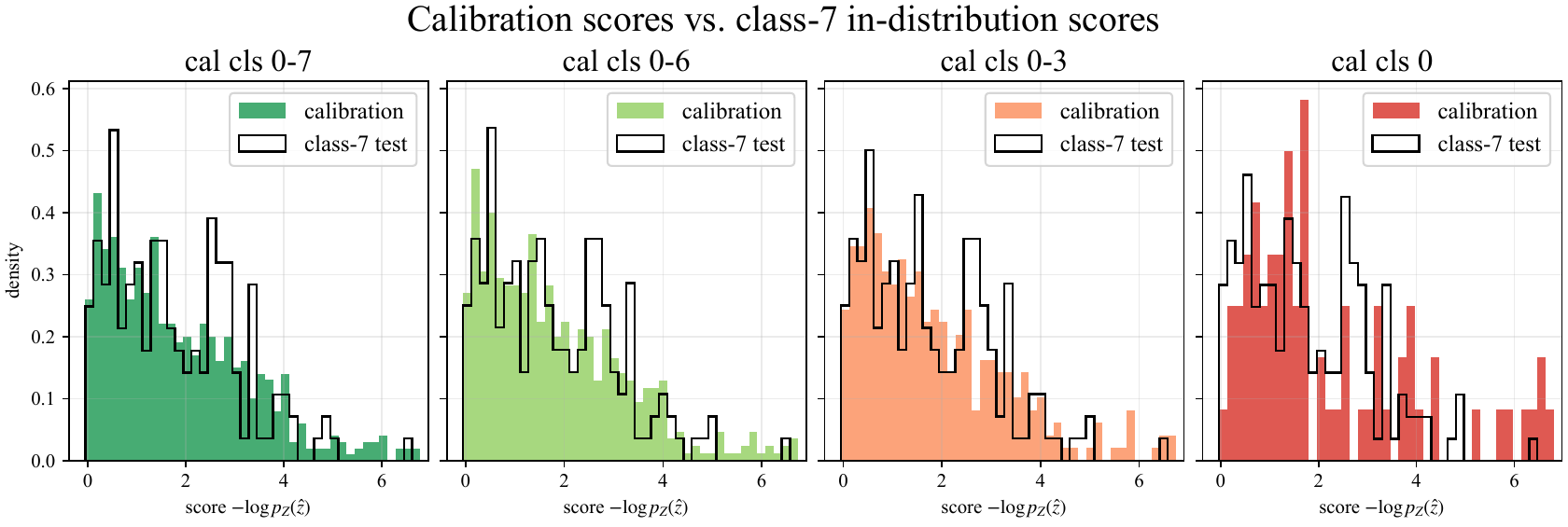}
    \caption{Calibration and true class-7 score distributions under
    reduced calibration coverage, using one reverse midpoint step
    (top) and five steps (bottom). Class-7 test scores overlap the
    calibration distributions in the displayed coverage settings.}
    \label{fig:appendix-ID-hist}
\end{figure}

Figure~\ref{fig:appendix-OOD-hist} evaluates points from classes 6, 5, and 0 under the incorrect class-7 condition, with calibration limited to classes 0--6. Their scores shift away from the reference distributions, and all three displayed wrong-label groups have $0.0\%$ acceptance under both integration budgets. Together, the paired figures illustrate conditional discrimination with the one-step score
and its five-step counterpart in this controlled setting. Both integration budgets reject all displayed wrong-label groups and produce qualitatively similar class-specific acceptance regions.

\begin{figure}[t]
    \centering
    {\small\textbf{One step}\par}
    \includegraphics[width=\linewidth]{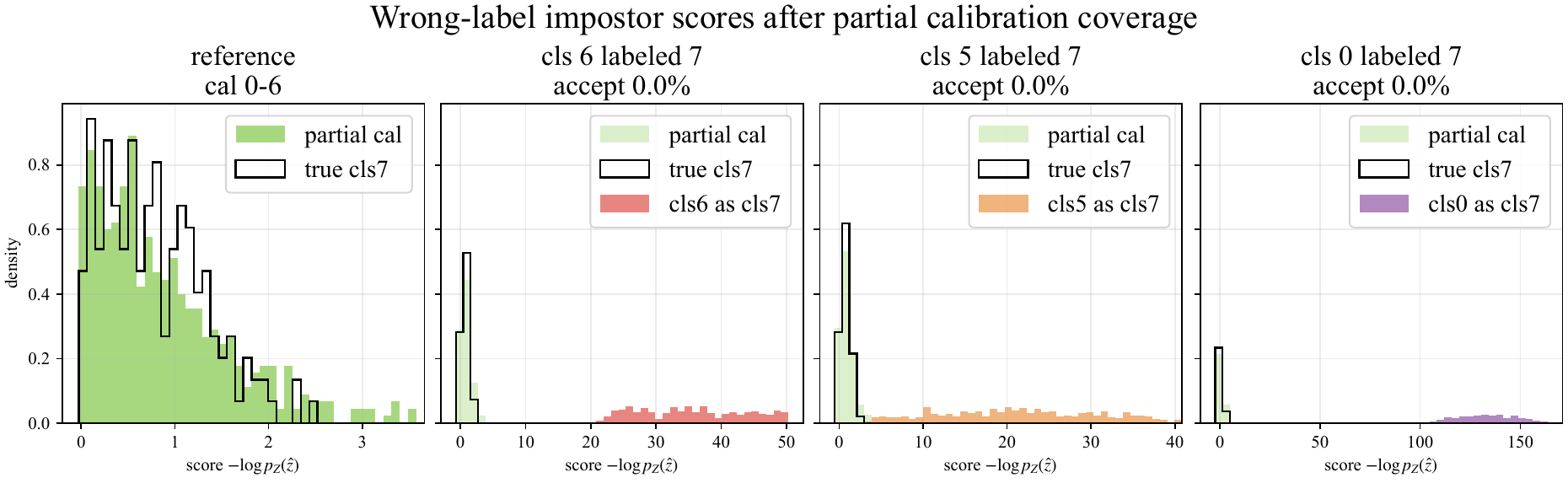}
    \medskip
    {\small\textbf{Five steps}\par}
    \includegraphics[width=\linewidth]{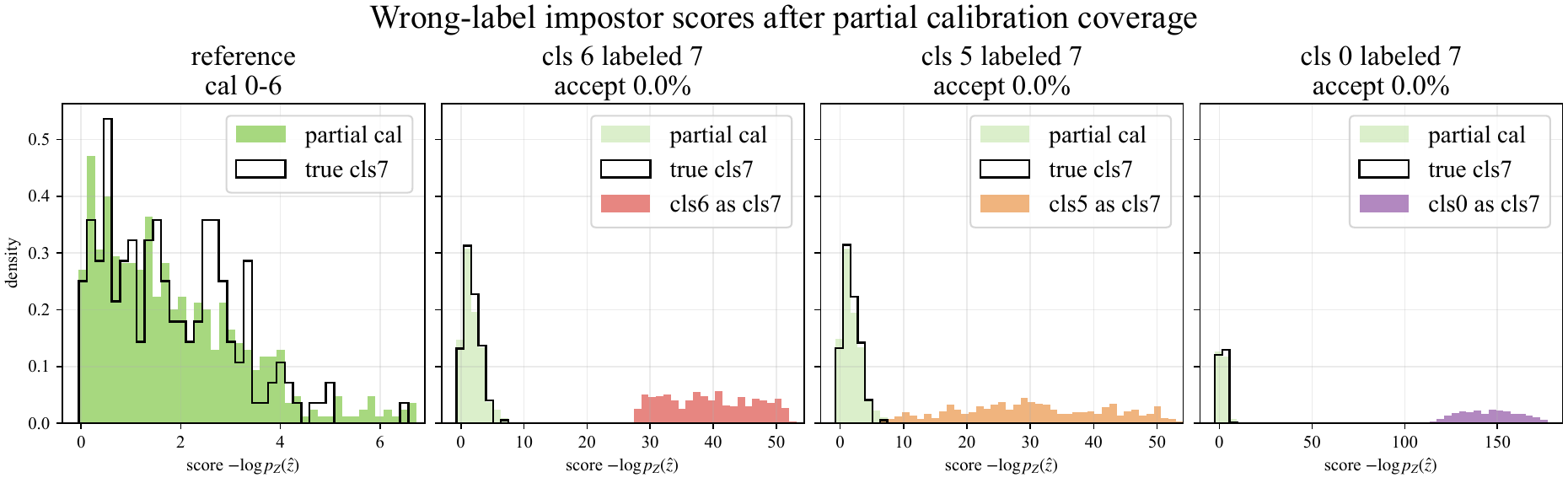}
    \caption{Mismatched class conditions with one reverse midpoint step
    (top) and five steps (bottom). Calibration uses classes 0--6; samples
    from classes 6, 5, and 0 are queried as class 7. Each displayed
    wrong-label group has $0.0\%$ acceptance at $\alpha=0.10$.}
    \label{fig:appendix-OOD-hist}
\end{figure}

\needspace{18\baselineskip}
\subsection{Auxiliary validation through score-guided local search.}
\label{parag: local search}

\begin{wrapfigure}[16]{r}{0.33\textwidth}
    \vspace{-\baselineskip}
    \centering
    \includegraphics[width=\linewidth]{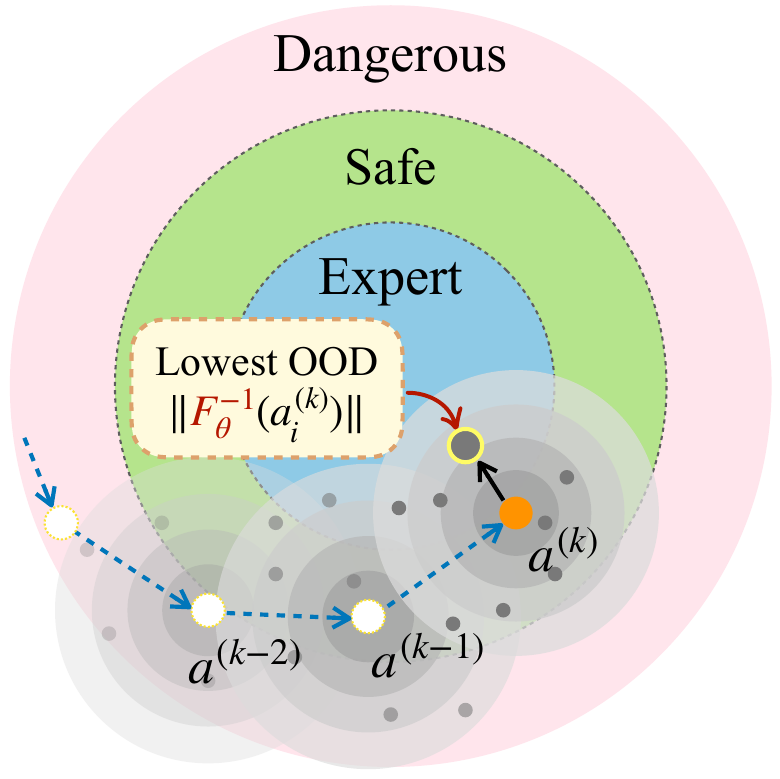}
    \caption{\small Score-guided local search.}
    \label{fig:score-guided-local-search}
    \vspace{-\baselineskip}
\end{wrapfigure}

We further evaluate the effectiveness of the OOD score through an auxiliary score-guided local search experiment. Specifically, we test whether using the score to guide local changes to action chunks from the agent's distribution
improves downstream task success (Table~\ref{tab:ood_local_search}).\footnote{A low score suggests consistency with expert data but does not necessarily imply that an action is useful for the current task. This search procedure is used to evaluate the score and is not a component of the \alg assistance pipeline.}

As illustrated in Figure~\ref{fig:score-guided-local-search}, we initialize the search at the agent's proposal, $a^{(0)}=a_u$, and evaluate candidates using
\begin{equation}
    \ell_\tau(a;I,a_u)
    = \max\!\left(s(a;I)-\tau,0\right)
    + \lambda_{\mathrm{align}}\|a-a_u\|_2^2,
    \label{eq:ood_search_objective}
\end{equation}
where $\tau$ is the calibrated score threshold corresponding to the gating level $\alpha$.
The loss penalizes scores above $\tau$ and large deviations ($\lambda_{\mathrm{align}}\geq0$) from the original proposal.

At each iteration $k=0,\ldots,K-1$, we draw $M$ local
perturbations and include a zero perturbation:
\begin{equation}
    \epsilon_i^{(k)}
    \sim \mathcal{N}(0,\sigma^2 I),
    \quad i=1,\ldots,M,
    \qquad
    \epsilon_0^{(k)}=0.
    \label{eq:ood_search_perturbations}
\end{equation}
The candidate chunks are
\begin{equation}
    a_i^{(k)}
    =
    a^{(k)}+\epsilon_i^{(k)},
    \qquad i=0,\ldots,M.
    \label{eq:ood_search_candidates}
\end{equation}
The zero perturbation retains the current chunk as a candidate.
We evaluate
$\ell_i^{(k)}=\ell_\tau(a_i^{(k)};I,a_u)$
and assign larger weights to candidates with smaller losses:
\begin{equation}
    w_i^{(k)}
    =
    \frac{
        \exp\!\left[-\left(
            \ell_i^{(k)}-\ell_{\min}^{(k)}
        \right)\right]
    }{
        \sum_{j=0}^{M}
        \exp\!\left[-\left(
            \ell_j^{(k)}-\ell_{\min}^{(k)}
        \right)\right]
    },
    \qquad
    \ell_{\min}^{(k)}
    =
    \min_{0\leq j\leq M}\ell_j^{(k)}.
    \label{eq:ood_search_weights}
\end{equation}
Subtracting the minimum loss provides numerical stability without changing the normalized weights. We then update the current chunk using the weighted perturbation:
\begin{equation}
    a^{(k+1)}
    =
    a^{(k)}
    +
    \eta_{\mathrm{edit}}
    \sum_{i=0}^{M}
    w_i^{(k)}\epsilon_i^{(k)},
    \label{eq:ood_search_update}
\end{equation}
where $\eta_{\mathrm{edit}}>0$ controls the update magnitude. The regularization remains anchored at the original proposal $a_u$ throughout the search. After $K$ iterations, we execute the refined chunk $\tilde a=a^{(K)}$.

\paragraph{Setup.}
We run the search on the goal-conditioned IL agent of Section~\ref{sec:controlled_results}, under the four perturbations Noised, Laggy, Slow, and Shift with the parameters in Appendix Table~\ref{tab:wrapper_params}. Evaluation uses the same episodes as the other methods in that section, so the Unassisted column matches Table~\ref{tab:task_success_summary_transposed}. The search uses only the expert flow and the agent's proposal. It needs no goal, no task reward, and no additional training. For each task, we average success equally over the four perturbations.

\paragraph{Results.}
Table~\ref{tab:ood_local_search} shows that the search improves average success on all three tasks. The gain is largest on Keypad, from $48.5\%$ to $65.0\%$, and smallest on CAN, from $44.2\%$ to $46.5\%$. The search changes a chunk only to lower its score while keeping it close to the proposal, so these gains indicate that the score gives a useful local direction for correcting perturbed actions. The search is also expensive. It scores every candidate at each iteration and takes $1.68$~s per query, about $22$ times the FLOPs of a FLOWER query (Table~\ref{tab:inference_cost}). We therefore use it only to validate the score, not as an editor in \alg.

\begin{table}[!ht]
    \centering
    \caption{Success rate (\%) of the perturbed IL agent with and without score-guided local search, averaged over the four perturbations.}
    \label{tab:ood_local_search}
    \small
    \begin{tabular}{lccc}
        \toprule
        Task & Unassisted & Local search & $\Delta$ \\
        \midrule
        Slalom & $40.6$ & $48.7 \pm 3.63$ & $+8.1$ \\
        CAN    & $44.2$ & $46.5 \pm 2.70$ & $+2.3$ \\
        Keypad & $48.5$ & $65.0 \pm 2.38$ & $+16.5$ \\
        \bottomrule
    \end{tabular}
\end{table}

\section{Computational Cost}
\label{app:compute_cost}

\paragraph{Measurement setup.}
Table~\ref{tab:inference_cost} profiles one adapter query $(I,a_u)\mapsto\tilde a$ at batch size one on an NVIDIA RTX A6000 with $16$ AMD CPU cores. A query takes the image observation and the agent's proposed chunk and returns the refined chunk. GFLOPs/query counts both the conditioning and the velocity evaluations. Peak memory is the largest memory use during one query. \acs{VLA}-query equiv divides GFLOPs/query by the cost of one FLOWER query, $888.15$ GFLOPs. Median latency is measured per task, and the table reports the summary of these task-wise medians.

\paragraph{Call counts.}
Calls/query counts image encodings plus velocity-field evaluations; a batched call counts once. \acs{GRIP} is profiled at $t_{\mathrm{init}}=0.8$ on a $100$-step grid, so it integrates $20$ intervals. Each interval is a predictor--corrector step with three velocity evaluations, which gives $60$ velocity evaluations and $2$ image encodings. The energy gradient acts only on the alignment term and adds no network calls. \acs{LIFT} runs $K=50$ editing steps and evaluates both flows at each step, which gives $100$ velocity evaluations and $4$ image encodings. Local search runs $K=15$ iterations. Each iteration encodes the image once and scores all $M+1=257$ candidates with one midpoint reverse step in a single batch, which gives $15$ encodings and $30$ velocity evaluations.

\paragraph{Results.}
\acs{GRIP} costs $2.5\%$ of a FLOWER query in FLOPs and takes $153.60$ ms. \acs{LIFT} costs $17.8\%$ and takes $202.11$ ms. \acs{LIFT} is more expensive because it evaluates both the agent flow and the expert flow at each editing step, while \acs{GRIP} uses only the expert flow and integrates from $t_{\mathrm{init}}$ to $1$. One action-chunk correction therefore takes about $0.15$--$0.20$ s in this benchmark. The cost of \acs{GRIP} grows with the number of integration intervals: on Slalom, its latency ranges from $34$ ms at $t_{\mathrm{init}}=0.95$ to $323$ ms at $t_{\mathrm{init}}=0.5$. Score-guided local search takes $1.68$ s per query and $2192\%$ of a FLOWER query in FLOPs, because it scores $M+1$ candidates at each of its $K$ iterations. We use it only to test whether the OOD score is a useful signal (Appendix~\ref{parag: local search}). It is not part of \alg.

\begin{table}[!ht]
    \centering
    \scriptsize
    \setlength{\tabcolsep}{3pt}
    \caption{Batch-one adapter inference on an RTX A6000.
    Latency summarizes task-wise medians; FLOPs include conditioning
    and velocity evaluations. \acs{VLA}-query equiv is relative to
    one FLOWER query ($888.15$ GFLOPs).}
    \label{tab:inference_cost}
    \begin{tabular*}{\textwidth}{@{\extracolsep{\fill}}lrrrrr@{}}
    \toprule
    Method & Calls/query & GFLOPs/query & Peak memory & \acs{VLA}-query equiv & Median latency (ms) \\
    \midrule
    \acs{GRIP}      & 62  & $22.1$     & $43$ MiB      & $2.5\%$  & $153.60$ \\
    \acs{LIFT}        & 104 & $157.9$    & $527$ MiB     & $17.8\%$ & $202.11$ \\
    Local search (aux.) & 45  & $19{,}466$ & $2{,}768$ MiB & $2192\%$ & $1{,}682.8$ \\
    \bottomrule
    \end{tabular*}
\end{table}

\end{document}